\pdfoutput=1
\documentclass[11pt,a4paper]{article}
\usepackage[utf8]{inputenc}
\usepackage[T1]{fontenc}
\usepackage{times,latexsym}
\usepackage[round]{natbib}
\usepackage[acceptedWithA,nohyperref]{tacl2021v1}
\usepackage{amsmath,amssymb,graphicx,booktabs,array,tabularx,enumitem,microtype,needspace,xspace,capt-of}
\graphicspath{{figures/}}
\usepackage{adjustbox,placeins}
\newcommand{\inc}{\mbox{\textsc{inconsistent}}\xspace}
\newcommand{\und}{\mbox{\textsc{underdetermined}}\xspace}
\newcommand{\tru}{\mbox{\textsc{true}}\xspace}
\newcommand{\fal}{\mbox{\textsc{false}}\xspace}
\title{Phantom Paradoxes: Language Models Treat Unresolved Circular Truth Dependence as Contradiction}
\author{Ji Ho Bae \\ JRTI \\ \texttt{jihobae@snu.ac.kr}}
\date{}
\newcommand{\CnfCI}{[30.4, 42.2]}
\newcommand{\CnfEst}{+36.4}
\newcommand{\CnfP}{$2\times10^{-5}$}
\newcommand{\GemCI}{[48.2, 66.1]}
\newcommand{\GemEst}{+57.1}
\newcommand{\GemP}{$4\times10^{-5}$}
\newcommand{\IHAP}{6\times10^{-5}}
\newcommand{\IHRCI}{[0.16, 0.47]}
\newcommand{\IHRCnfCI}{[$-$0.04, 0.23]}
\newcommand{\IHRCnfEst}{+0.09}
\newcommand{\IHRCnfP}{0.209}
\newcommand{\IHREst}{+0.31}
\newcommand{\IHRP}{$2.6\times10^{-4}$}
\newcommand{\IHUP}{0.733}
\newcommand{\LlaCI}{[$-$0.9, 24.1]}
\newcommand{\LlaEst}{+11.6}
\newcommand{\LlaP}{$0.092$}
\newcommand{\LomoGem}{0.05}
\newcommand{\LomoLla}{0.32}
\newcommand{\LomoQwe}{0.96}
\newcommand{\LomoSev}{0.98}
\newcommand{\MoneCI}{[29.3, 44.9]}
\newcommand{\MoneEst}{+37.1}
\newcommand{\MoneP}{$<10^{-5}$}
\newcommand{\NLCI}{[12.5, 32.8]}
\newcommand{\NLEst}{+23.4}
\newcommand{\NLP}{$0.0022$}
\newcommand{\PrimCI}{[25.9, 36.2]}
\newcommand{\PrimEst}{+31.0}
\newcommand{\PrimP}{$2\times10^{-5}$}
\newcommand{\QweCI}{[5.4, 18.8]}
\newcommand{\QweEst}{+11.6}
\newcommand{\QweP}{$0.0036$}
\newcommand{\SevCI}{[33.0, 54.5]}
\newcommand{\SevEst}{+43.8}
\newcommand{\SevP}{$4\times10^{-5}$}
\newcommand{\SensCVGem}{0.99}
\newcommand{\SensCVLla}{1.00}
\newcommand{\SensCVSev}{1.00}
\newcommand{\SensCVQwe}{1.00}
\newcommand{\SensLomoGem}{0.62}
\newcommand{\SensLomoLla}{0.69}
\newcommand{\SensLomoSev}{0.85}
\newcommand{\SensLomoQwe}{0.80}
\newcommand{\JointGem}{0.64}
\newcommand{\JointLla}{0.70}
\newcommand{\JointSev}{0.88}
\newcommand{\JointQwe}{0.90}
\newcommand{\JointHolmGem}{0.360}
\newcommand{\JointHolmLla}{0.360}
\newcommand{\JointHolmSev}{0.008}
\newcommand{\JointHolmQwe}{0.030}
\newcommand{\IHRRoundOneEst}{+0.61}
\newcommand{\IHRRoundOneCI}{[0.38, 0.86]}
\newcommand{\IHRRoundOneP}{3\times10^{-5}}
\newcommand{\IHRRoundMultiEst}{+0.03}
\newcommand{\IHRRoundMultiCI}{[$-$0.11, 0.18]}
\newcommand{\IHRRoundMultiP}{0.697}
\begin{document}
\maketitle
\begin{abstract}
Are language models vulnerable to self-reference itself or to structures in which following references step by step does not settle the truth value? Our findings support the latter. We construct systems of utterances that circularly refer to one another's truth values but admit exactly one consistent truth-value assignment. Following the references step by step (propagation) cannot settle the queried utterance's truth value, whereas searching through all possible cases yields the unique solution. Thus, answering \inc{}, meaning that no consistent assignment exists, is always incorrect. We call this error a \emph{phantom paradox}: language models mistake satisfiable systems for contradictions and report nonexistent paradoxes. Across four open language models from three families, the rate of \inc{} answers on circular systems is, on average, 31.0 percentage points higher than on grounded twins, which preserve the correct answers but remove only the reference cycles; 36.4 percentage points higher than on constraint twins, which make no mention of truth; and 37.1 percentage points higher than on perplexity-matched controls. In the models' internal representations, the queried utterance's truth value is recovered in grounded twins and perplexity-matched controls but not in circular systems. This difference between grounded twins and circular systems appears in all four models. The truth value is also not recovered in constraint twins, yet \inc{} answers are much rarer there. A post hoc analysis further shows that recovery in grounded twins occurs mainly in graph families in which a single round of propagation settles the queried utterance. Recovery is weak and inconsistent in families requiring multiple rounds, yet the difference in the rate of \inc{} answers is the same in both groups (31.0 percentage points each). Therefore, failure to recover the truth value cannot by itself explain phantom paradoxes. It is reasonable to infer that they arise from a structure in which truth values depend on truth values that propagation cannot settle.
\end{abstract}

\section{Introduction}

Judging whether a set of statements can hold together is a basic task in logical reasoning, and benchmarks routinely evaluate this ability in language models. Deductive benchmarks test reasoning from premises to conclusions (\citealp{clark2020transformers}; \citealp{tafjord2021proofwriter}; \citealp{saparov2023greedy}; \citealp{han2022folio}), constraint satisfaction benchmarks test the ability to find consistent assignments (\citealp{lin2025zebralogic}; \citealp{wei2025satbench}), and paradox benchmarks test whether models can identify and explain the contradiction in a paradox (\citealp{chen2026parasuite}). We address the opposite error, which we call a \emph{phantom paradox}: reporting a contradiction in statements that can hold together.

The statement ``This sentence is false'' has no consistent truth value. The statement ``This sentence is true'', by contrast, admits two consistent truth-value assignments. The system ``A says that B is true, B says that C is false, and C says that A is false and B is true'' appears similarly tangled, but it has a unique consistent truth-value assignment (A and B are true, and C is false).

The truth values in the example above cannot, however, be settled by following the references step by step, a procedure we call \emph{propagation} (Kripke's least fixed-point construction; \citealp{kripke1975outline}). A's value can be settled only after B's, B's only after C's, and C's in turn only after that of A or B. An exhaustive search over all eight possible assignments nevertheless finds the consistent solution. Similarly, repeatedly substituting a specific number for an unknown in an equation may continue indefinitely even though a particular value is a solution.

The closest existing experiments involve knights-and-knaves puzzles, in which speakers make statements about who is telling the truth, and it is already well known that language models do not solve these puzzles reliably (\citealp{mondorf2024truthquest}; \citealp{xie2024memorization}). However, because all the puzzles in these studies contain a cycle of truth dependence while their difficulty varies with the number of speakers or the form of the statements, it is difficult to separate failures caused by circularity from failures caused by difficulty.

We show that language models have difficulty processing statements of this kind and ask whether the cause is self-reference itself or a structure in which propagating truth values does not settle the answer. Our hypothesis is the latter: models do not err because of self-reference alone. Rather, they treat truth recursion that propagation alone cannot settle (recursion unresolved at finite depth; see §2) as a signal of contradiction. Circular systems like the example above have a unique solution, yet their values cannot be settled step by step. If our hypothesis is correct, models should therefore issue contradiction verdicts more often on such systems than on systems that preserve every utterance's truth value but remove only the cycles.

Testing this hypothesis is not straightforward. Self-referential problems are longer and more awkwardly phrased than ordinary problems, leaving open a simpler explanation: language models may merely react differently to unfamiliar inputs (\citealp{gonen2023demystifying}; \citealp{mccoy2024embers}). An overly powerful probe can also learn the task on its own, so we must be clear about what a linear readout of a nonlinear network measures (\citealp{hewitt2019control}; \citealp{belinkov2022probing}). We read internal states only along truth directions validated layer by layer. A truth direction is a direction in a language model's internal state space that separates true from false statements (\citealp{marks2024geometry}).

For our primary analysis, we selected only circular systems that have exactly one consistent truth-value assignment. For these systems, the verdict \inc{}, meaning that no consistent assignment exists, is always incorrect. To determine whether circularity causes these errors, we built two twins for each circular system. A grounded twin keeps every truth value and the correct answer but breaks only the cycles, allowing propagation to settle its values within finitely many steps (Figure~1). A constraint twin keeps the solution set but states the conditions as formulas without a truth predicate. To rule out the unfamiliar-input explanation, we also selected perplexity-matched controls, grounded variants whose likelihood under each model is similar to that of the circular system. We formulated this hypothesis in a pilot experiment and tested it on four open models from three families using a separately frozen confirmatory dataset. The testing procedures were prespecified before we examined the target data for each analysis (§4).

Three results bear out our hypothesis. Circular systems draw far more \inc{} verdicts than their grounded twins (31.0 percentage points, with similar effects in the true/false and on/off wordings), constraint twins (36.4 percentage points), and perplexity-matched controls (37.1 percentage points). In natural-language items, ungrounded self-reference, whose value is never settled by propagation (\citealp{kripke1975outline}), likewise draws more \inc{} verdicts than grounded self-reference with the same classical value.

Inside the models, the state at the question's last token, read along the validated truth directions, leans further toward the correct answer in grounded twins than in circular systems (0.31 standard deviations, significant in all four models). In circular systems, it separates \tru{} from \fal{} correct answers only at chance level. Two further predictions, about late-layer revision and attention spread (P3 and P4; §2.5), are not supported (§6.2).

The difference along the truth directions appears only in graph families in which the grounded twin settles the queried utterance in the first round. Yet the gap in the rate of \inc{} answers is 31.0 percentage points in both groups of families. Moreover, constraint twins draw far fewer \inc{} verdicts even though their truth value is not recovered either. We conclude that failure to recover the truth value is not enough on its own: phantom paradoxes arise from a structure in which truth values depend on truth values that propagation cannot settle.

\setcounter{figure}{0}
\begin{figure}[t]
\centering
\includegraphics[width=\columnwidth]{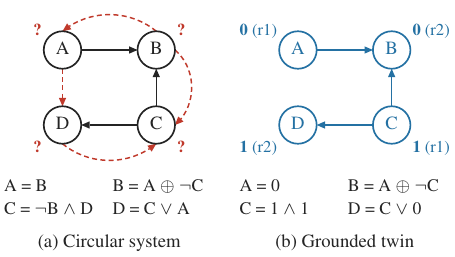}
\caption{Overview. (a) A circular system: an edge $j\to i$ means that utterance $i$ uses $j$'s value; dashed edges are cycle-closing references (order C, D, A, B; §2). (b) Its grounded twin replaces them with the values they take under the unique consistent assignment \mbox{(A, B, C, D) = (0, 0, 1, 1)}; $\oplus $ is exclusive OR. Labels show the result of propagation from ``unknown'' (§2): every value in the circular system remains unknown (?), whereas each value in the grounded twin is settled in the round shown, with the queried utterance D \tru{}.}
\end{figure}

\section{Truth Recursion and Finite-Depth Resolution}

\subsection{Systems and consistency}
Each system consists of $n$ utterances, each identified by a distinct name. Utterance $i$ asserts a formula $F_i$ that combines constants and utterances' truth values using Boolean operations (for example, ``Utterance B is false and utterance D is true''). The formula may also include utterance $i$'s own value. We define \emph{truth recursion} as evaluating an utterance's value from the values of the utterances it mentions. For $F=(F_1,\ldots ,F_n)$, a truth-value assignment $x\in \{0,1\}^n$ is a consistent assignment if it satisfies $x=F(x)$. Let $\mathcal{S}=\{x:x=F(x)\}$ be the set of consistent assignments, and let $V_q=\{x_q:x\in \mathcal{S}\}$ be the set of values assigned to the queried utterance $q$ by those assignments. The correct answer is labeled \inc{} if $\mathcal{S}=\emptyset $, \tru{} if $V_q=\{1\}$, \fal{} if $V_q=\{0\}$, and \und{} if $V_q=\{0,1\}$.

Circularity means that some utterance's truth value eventually depends on itself through references. That is, when utterance $i$ in a system mentions the truth value of utterance $j$, we represent that dependence with an edge $j\to i$ in the dependency graph. If the graph has at least one cycle, we call the system circular; otherwise, it is acyclic. A circular system may have no consistent assignment (the liar; a paradox), multiple consistent assignments (the truth-teller; a hypodox; \citealp{rabern2013dangerous}), or exactly one.

Consistency concerns whether a solution exists. If there is at least one consistent assignment, the system is consistent; if there is none, it is inconsistent (\inc{}). Our study focuses on circular systems with exactly one consistent assignment, which are consistent despite their circularity.

\subsection{Finite-depth resolution}
Propagation, introduced above, is the evaluation that settles values in turn, starting from utterances whose values are already settled. Formally, it is defined by setting all utterances to unknown and, in each round, evaluating all $F_i$ simultaneously under strong Kleene logic, using the values from the preceding round.

In the finite systems considered here, the values settled within a finite number of rounds form the least fixed point of \citet{kripke1975outline}. If the queried utterance has a value at this fixed point, we say that the truth recursion is resolved at finite depth (grounded in Kripke's sense); otherwise, we say that it is unresolved (ungrounded). In an acyclic system, all values are necessarily settled within at most $n$ rounds. By contrast, in a circular system without constants, no value is ever settled, whether the system has zero, one, or several consistent assignments. Finding the answer then requires solving $x = F(x)$ as a whole (\citealp{walicki2009reference}).

In Figure~1, the grounded twin resolves in two rounds. In the first round, A takes the constant value \fal{} and C (``\tru{} and \tru{}'') is settled. In the second round, B (``A is true or C is false, but not both'') and D (``C is true or \fal{}'') are settled, and the queried utterance D is \tru{}. In the circular system, by contrast, every utterance waits for another, so evaluation makes no progress from the start and every value remains unknown, even though there is exactly one consistent assignment.

\subsection{Grounded and constraint twins}
A grounded twin is obtained from a circular system by replacing all of its \emph{cycle-closing references} with constants. When the utterances are arranged in a random order independent of their presentation order, a cycle-closing reference is one that points to the utterance containing it or to a later utterance. The replacement constant is the reference's value in the unique consistent assignment (``B is false'' becomes \tru{} if B is false). The two systems therefore share the same consistent assignment, correct answer, and remaining wording, whereas the grounded twin's dependency graph has no cycles (Figure~1). Exact computation confirms that truth recursion is unresolved at finite depth in all core circular systems (those with exactly one consistent assignment) and resolved in all of their grounded twins.

A \emph{constraint twin} expresses the same Boolean conditions as the circular system in conjunctive normal form (CNF) over named variables and does not include utterances about truth. The constraint twin preserves the variables and solution set exactly but eliminates the form of truth recursion.

\subsection{Two notions of resolution}
Finite-depth resolution asks which truth values are settled starting from the ``unknown'' state. Revision theory (\citealp{gupta1993revision}) asks a different question: whether repeated synchronous reevaluation (reevaluating all utterances simultaneously) eventually gives an utterance the same stable value regardless of the initial assignment. We use the strictest criterion: whether all initial assignments converge to the same state. We represent one step of this reevaluation by the $2^n\times 2^n$ matrix $P$ ($Pe_x=e_{F(x)}$) that sends the basis vector $e_x$ of each assignment $x$ to the basis vector $e_{F(x)}$ of the new assignment. Let $Q$ be the diagonal matrix whose diagonal entries are $1$ for assignments with $x_q=1$ and $0$ otherwise. Since a consistent assignment is a fixed point of one step of reevaluation (i.e., a self-loop of $P$), $\mathrm{tr}(QP)$ and $\mathrm{tr}((I-Q)P)$ represent the numbers of consistent assignments in which $q$ is true and false, respectively, and these two values determine the correct answer label.

In an acyclic system, $P^n$ sends all initial assignments to the same state, so $\mathrm{rank}(P^n)=1$. However, such convergence does not imply acyclicity. Even a circular system can satisfy $\mathrm{rank}(P^k)=1$ for some $k$. We found that 15 of the 56 core circular systems (those with exactly one consistent assignment; §3) are completely unresolved at finite depth yet exhibit such convergence. Because these two notions of resolution diverge, we compare in §7 which one language models follow.

\subsection{Predictions}
We present five predictions based on two assumptions: first, that language models process truth recursion like the propagation defined above, that is, by successively following references and settling values in turn; and second, that they interpret an unresolved query as a sign of contradiction. P1 is the central behavioral prediction and P2 its internal counterpart. P3 and P4 introduce an additional assumption about how the network processes an unresolved item and can therefore fail independently of P1 and P2.
\begin{description}[leftmargin=2.6em,labelwidth=2.2em,labelsep=0.4em,itemsep=2pt,topsep=4pt,font=\normalfont\bfseries]
\item[{[P1]}] \textbf{Phantom paradoxes.} Satisfiable circular systems receive \inc{} verdicts more often than their corresponding grounded twins.

\item[{[P2]}] \textbf{Reduced recovery of the truth value.} At the question's last token, recovery of the queried value, measured along the validated truth direction, is lower in circular systems than in their grounded twins.

\item[{[P3]}] \textbf{Continued revision.} At the question's last token, the magnitude of the late-layer revision to the internal state (late-layer update) is larger in circular systems than in their grounded twins.

\item[{[P4]}] \textbf{Attention spread.} Attention from the last 32 tokens of the prompt spreads over more effective dimensions in circular systems than in their grounded twins.

\item[{[P5]}] \textbf{Specificity.} For the measures of P2--P4, constraint twins show the same pattern as grounded twins.
\end{description}

The formalism concerns Boolean assignments, whereas attention matrices act on token positions and residual states pass through normalization and nonlinear maps. We do not assume here that the network implements $P$ or that its layers are linear.

\section{Data}

\paragraph{Graph families.}
Our dataset consists of 144 dependency-graph families. Each graph has a designated queried utterance, which we call its \emph{root}. The systems in each family consist of four, five, or six named utterances (48 families of each size). We constructed this dataset before generating any responses. The circular systems are evenly distributed across the four correct answers (36 each of \tru{}, \fal{}, \und{}, and \inc{}). From these, we selected as our core set the 56 families in which the circular system has exactly one consistent assignment and the grounded twin preserves that entire assignment (31 \tru{} and 25 \fal{}). We also used a solver to identify these 56 eligible families before generating any responses.

\paragraph{Wordings, twins, and controls.}
Each circular system and its grounded twin are presented in two wordings. The \emph{truth wording} presents utterances that are true or false, and the \emph{bit wording} presents registers that are on or off. Each family contains the circular system and its grounded twin, each in both wordings, plus a constraint twin (we verified agreement of the solution sets independently by exhaustive enumeration and by Z3; \citealp{demoura2008z3}). As a separate reasoning baseline, we constructed 32 families of ordinary constraint problems with no self-reference at all. This baseline was not used in any contrast (results in Table~4 in Appendix~D).

\paragraph{Perplexity-matched controls.}
The results on circular systems could simply reflect the lower frequency of their text relative to that of their twins. To test this, we constructed 13 \emph{grounded variants} per family that differed only in the utterance names. To vary textual likelihood without changing the logical structure, we used the usual single letters and random strings of one to four letters as utterance names.

For each model, we computed the negative log-likelihood (NLL) of each prompt body and of the entire user prompt (content). For each input row presenting a circular system in the truth wording, we then searched for a grounded variant that satisfied all three conditions: $|\Delta \mathrm{NLL}_{\mathrm{body}}| \leq 0.35$ nats, $|\Delta \mathrm{NLL}_{\mathrm{content}}| \leq 0.35$ nats, and $|\mathrm{log}(T_{\mathrm{src}}/T_{\mathrm{var}})| \leq 0.15$. Here, $T$ is the number of tokens in the body (src: circular system; var: candidate variant), and NLL is the mean negative log-likelihood per token. Of the variants that met these conditions, we chose as the match the one closest in body likelihood and length (see Appendix~A for the exact rule and the two likelihood spans).

\paragraph{Prompts and answers.}
Every prompt presents the four-way classification rule, which includes the statement ``INCONSISTENT if no simultaneous assignment satisfies all conditions'' (full text in Appendix~B). The prompt then presents the system and asks for brief reasoning followed by a final verdict line (``FINAL: LABEL''). We treat a response as parsed only if it contains exactly one line of this form and ends with that line. All other responses count as unparsed but remain in every denominator. We also report adversarial bounds, defined in §4, obtained by scoring unparsed responses adversarially.

\paragraph{Calibration propositions and natural-language items.}
For the internal analysis, truth directions are fitted to 128 arithmetic propositions independent of the test items (64 true and 64 false; §6). These propositions use the same prompt format as the test items (e.g., the false proposition ``the sum of 35 and 34 equals 70'').

We also built a separate natural-language dataset of 64 items, combining each of 16 wording families with four forms of self-reference: truth-teller, liar, grounded self-reference, and classically determinate but ungrounded self-reference. Circularity and groundedness are not separated in the main dataset, so these items test whether Kripkean groundedness (\citealp{kripke1975outline}), rather than circularity, governs the verdicts. Within each determinate pair, the classical solution, word count, and punctuation are identical.

\section{Experimental Setup}

\paragraph{Origin of the hypothesis.}
We formulated our hypothesis through a development pilot experiment on four graph families constructed outside the dataset. For two circular systems with a unique solution, 10 of 16 responses were \inc{}, compared with 3 of 16 after cycle removal.

We fixed the hypotheses and analysis methods before examining the target data for each analysis. P1 was fixed before the first confirmatory response was generated; P2--P5 were fixed after we examined Gemma's responses and checked the output text of Llama-70B's first batch for technical defects, but before we read any hidden states or attention values of the confirmatory dataset. These tests exclude analyses outside the original plans, including the larger batch-size-one rerun in §5; the recomputation of the update without the final step, the split by rounds, and the training-row scaling check in §6; the comparison of converging and nonconverging circular systems in §7; and the classifier restricted to the resolution score features in Appendix~C.

\paragraph{Models and decoding.}
The four open models we used are Gemma-2-9B-it (hereafter Gemma; \citealp{gemma2}), Llama-3.2-11B-Vision-Instruct (Llama-11B), Llama-3.3-70B-Instruct (Llama-70B), and Qwen3-VL-8B-Instruct (Qwen; \citealp{qwen3vl}). Both Llama models belong to the Llama 3 family (\citealp{llama3}). We provided text-only input to both vision-capable models (Llama-11B and Qwen). We ran all models using fixed repository revisions in BF16 without quantization (Appendix~A). We generated up to 4,096 new tokens with greedy decoding. We processed input rows in batches of 16 or 32 with left padding. We formed batches by input length, regardless of condition, before the first response was generated, and afterward recomputed hidden states by teacher forcing on the generated sequences (Appendix~A).

\paragraph{Endpoints and statistics.}
The primary endpoint is an indicator of whether the final answer is \inc{}. For each of the 56 core families, we averaged the difference (circular system minus grounded twin) across the two wordings (truth and bit) and then across the four models, with equal weights. This yielded one value per family.

We evaluated the mean difference using a family-level two-sided paired sign-flip randomization test (100,000 draws; \citealp{yeh2000accurate}; \citealp{dror2018hitchhiker}). We also report 95\% confidence intervals (CIs) obtained by family-level bootstrap resampling (10,000 draws; \citealp{koehn2004statistical}; \citealp{bergkirkpatrick2012empirical}). We applied Holm correction (\citealp{holm1979simple}; \citealp{dror2017replicability}) separately within each test family: the primary test and the behavioral specificity contrast (circular systems versus constraint twins), the internal tests P2--P4, the per-model tests of each contrast or measure across the four models, the 15 single-feature tests, and each classifier's transfers to the four held-out models. All other $p$-values are unadjusted. In the primary estimate, we treated unparsed responses as not \inc{}; to compute the adversarial upper and lower bounds, we scored them to maximize or minimize, respectively, the difference in the rate of \inc{} answers between the circular systems and the comparison items.

\paragraph{Data release.}
All prompts, raw outputs, full hidden states and attention views, the generation and analysis code, an exact solver, and both datasets with their generators and labels will be released with the paper, under an open license, on public research-data repositories with persistent identifiers.

\section{Behavioral Results}

\setcounter{table}{0}
\begin{table}[t]
\centering\footnotesize
\setlength{\tabcolsep}{3pt}
\begin{tabular}{@{}lrcr@{}}
\toprule
Contrast & pp & 95\% CI & $p$ \\
\midrule
Circular $-$ grounded twin & \PrimEst & \PrimCI & \PrimP \\
Circular $-$ constraint twin & \CnfEst & \CnfCI & \CnfP \\
Circular $-$ matched control & \MoneEst & \MoneCI & \MoneP \\
\addlinespace
\multicolumn{4}{@{}l}{\emph{Circular $-$ grounded twin, by model}}\\
\quad Gemma-2-9B & \GemEst & \GemCI & \GemP \\
\quad Llama-3.2-11B & \LlaEst & \LlaCI & \LlaP \\
\quad Llama-3.3-70B & \SevEst & \SevCI & \SevP \\
\quad Qwen3-VL-8B & \QweEst & \QweCI & \QweP \\
\addlinespace
\multicolumn{4}{@{}l}{\emph{Natural language (16 wording families)}}\\
\quad Ungrounded $-$ grounded & \NLEst & \NLCI & \NLP \\
\bottomrule
\end{tabular}
\caption{Differences in the rate of \inc{} answers (percentage points, 56 core families; averaged over wordings within families, then over models). The perplexity-matched contrast uses the truth wording and the models with a matched control. $p$: family-level sign-flip randomization test (§4), Holm-adjusted within the first two rows and within the per-model rows, unadjusted elsewhere.}
\label{tab:behavior}
\end{table}

\setcounter{figure}{1}
\begin{figure}[t]
\centering
\includegraphics[width=\columnwidth]{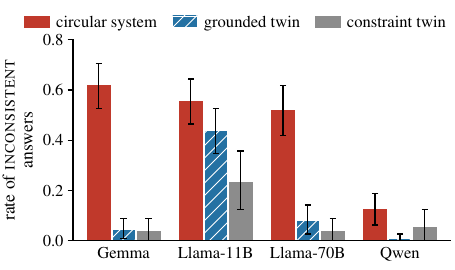}
\caption{Rate of \inc{} answers by model on the 56 core families (wordings averaged within each family; error bars are 95\% family-bootstrap intervals). Within each core family, the three item types share the same solution and correct answer.}
\end{figure}

\subsection{Circular systems versus grounded twins}
The primary hypothesis is confirmed. Across the four models (Table~1), the rate of \inc{} answers on the 56 core circular systems was, on average, 31.0 percentage points higher than on their grounded twins (Figure~2). Even when all unparsed responses, including rows truncated at the token limit (4.0\% of the study's 3,286 rows, mostly Qwen; Table~4), were scored adversarially, the difference ranged from 30.8 to 37.1 percentage points. All four models showed a positive effect, as predicted. The effect was largest in Gemma and was significant in three of the four models. Accuracy showed the opposite pattern: averaged across models, it was 32\% on circular systems and 72\% on grounded twins. The circular--grounded contrast was present in both wordings (truth wording 29.9 percentage points, bit wording 32.1 percentage points; difference $-$2.2, $p=0.751$), so the effect does not depend on the word ``true''.

\subsection{Perplexity-matched controls and constraint twins}
The effect persists after matching perplexity. Unusual inputs, rather than logic, might explain the effect, but all four models assigned higher likelihood per token to circular-system prompts than to grounded-twin prompts. Even against the perplexity-matched controls (§3), circular systems drew \inc{} verdicts 37.1 percentage points more often. Since some cases had no matched control, the comparison includes only the model--family combinations with a matched control (171 of 224; Table~3 in Appendix~D gives the numbers of matched prompts over all 144 families).

Among the core families with a matched control, the model-specific difference was +50.0 percentage points for Gemma (52 families), +15.7 for \mbox{Llama-11B} (51 families), +48.2 for Llama-70B (56 families), and +16.7 for Qwen (12 families). In 159 of these 171 model--family combinations, the selected control was the grounded twin itself: the variant with the usual single-letter names. In the remaining 12 cases, neither the control nor the grounded twin received an \inc{} answer. That is, in every matched case, the control and the grounded twin had the same rate of \inc{} answers.

\paragraph{Constraint twins.}
Equivalent constraints without truth recursion draw fewer \inc{} verdicts. Constraint twins (§3) have a unique solution but use longer prompts. Accuracy on these items was 18\%, lower than the 32\% on circular systems. Even so, circular systems drew \inc{} verdicts 36.4 percentage points more often than constraint twins (Table~1; adversarial bounds 26.1--42.4 percentage points). A constraint twin has the same solution set as its circular system, so the two do not differ logically. We can therefore conclude that the effect observed above depends not on the system's logic but on its presentation, specifically the form of circular reference.

\subsection{Secondary outcomes and robustness}
Secondary outcomes point in the same direction as the primary hypothesis. Across all 108 satisfiable circular systems, 43\% of responses were \inc{}. For their grounded twins, the rate was 16\% (see Table~4 for results by model). In the 56 core families, the proportion of incorrect or unparsed responses labeled \inc{} was 62\% on circular systems and 46\% on grounded twins (model average). In regressions adjusted for each row's body likelihood and length, circular systems also drew \inc{} verdicts 37.0 percentage points more often than their grounded twins (95\% CI [24.7, 48.4]), with a positive estimate in every model.

\paragraph{Groundedness in natural language.}
The same pattern appears in the natural-language dataset, which holds self-reference fixed and varies only groundedness (§3): classically determinate ungrounded items drew \inc{} verdicts 23.4 percentage points more often than grounded items with the same classical solution. The pattern was clearest in Qwen, which labeled every liar item and every ungrounded item with a false classical value \inc{} but answered every grounded item correctly.

However, the 16 wording families were not constructed with mutually independent logical structures. This dataset also does not separate ungrounded false items from contradictory content. When broken down by truth value, the difference was 37.5 percentage points for items whose consistent truth value is false and 9.4 percentage points for those whose value is true.

\paragraph{Robustness to batching.}
In the main experiment, rows of similar input length were processed in batches of 16 or 32. Under BF16 greedy decoding, a row's logits vary slightly with its batch, which can change the outcome of a long chain of reasoning. When 16 rows per model, chosen in advance by a hash of their identifiers and the same in all four models, were regenerated one at a time (batch size one), 41 of the 64 final labels were reproduced. The changes were not biased in favor of our prediction. After the main analysis, regenerating the four circular-system and grounded-twin rows from each of 16 core families, chosen by a hash of their identifiers and the same in all four models (64 rows per model), reproduced 175 of the 256 labels, and the contrast was +30.5 percentage points, close to the +27.3 under batched generation (difference between the unrounded estimates +3.1 percentage points, 95\% CI [$-$5.5, 10.2]; $p=0.586$).

\section{Internal Analysis}

\setcounter{figure}{2}
\begin{figure*}[t]
\centering
\includegraphics[width=\textwidth]{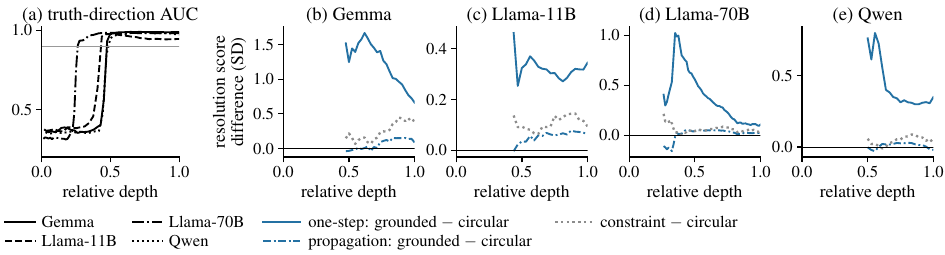}
\caption{Layerwise truth directions and resolution score differences at the question's last token. (a) AUC of each layer's truth direction on unseen arithmetic propositions (values below 0.5 are possible; Appendix~C); the gray line marks the 0.90 validation threshold. (b)--(e) Within-family differences in resolution score (comparison item minus circular system), averaged over families at validated layers (calibration standard deviations). Solid and dash-dotted lines: grounded twins in one-step (27) and propagation (29) families (§6.4); dotted line: constraint twins (truth wording). The vertical-axis ranges differ across panels.}
\end{figure*}

\setcounter{table}{1}
 Truth scores (defined in §6.1) and late-layer updates were computed from the residual stream at the question's last token (``Query: …''). The residual stream is the internal state to which each layer's output is added as the state passes through the layers. The late-layer update measures the relative magnitude of the change in this state across a single layer (the residual update), averaged over the last third of layer transitions (see Appendix~C for the definition). Attention rank is the effective rank, divided by 32, of the head-averaged outgoing attention from the prompt's last 32 tokens, all of which follow the question. Attention rank increases as these tokens attend in more distinct patterns, and P4 uses its value at middle depth. These tokens are identical in a circular system and its grounded twin.

\paragraph{Truth directions validated layer by layer.}
Because a truth direction obtained at one layer is not guaranteed to remain valid at other layers (\citealp{poulis2026limits}), we derived a separate truth direction for every layer of every model as the difference between the mean residual states of 64 true and 64 false arithmetic propositions. The classification threshold was placed midway between the two classes (mean-difference probe; \citealp{marks2024geometry}). A layer counts as validated when, in eightfold cross-validation, its direction reaches AUC $\geq 0.90$ on propositions not used to fit that direction. At every validated layer, the AUC was higher than any AUC obtained from 1,000 label permutations.

As Figure~3a shows, this AUC increases in deeper layers. The first layer to reach 0.90 is layer 20 of 42 in Gemma, 14 of 32 in Llama-11B, 21 of 80 in Llama-70B, and 18 of 36 in Qwen. All subsequent layers pass the threshold. In the truth-score analyses below, we use only layers that pass this threshold, and we summarize each model by the single best layer selected using the arithmetic calibration propositions alone (AUC 0.977--0.990). The value of the queried utterance is \emph{recovered} when its truth or falsity can be read out by projecting the internal state at the question's last token onto this truth direction.

\subsection{Recovery of the queried truth value (P2,~P5)}
At the best layer of each model, the \emph{truth score} is the projection of the internal state at the question's last token onto the truth direction, put on the same scale as the arithmetic propositions (Appendix~C); larger values lie further toward ``true''. Flipping the truth score's sign for systems whose correct answer is \fal{} gives the \emph{resolution score}; larger values lean further toward the correct answer. Since a circular system and its grounded twin have the same correct answer, the difference in their resolution scores shows which has the stronger signal toward the correct answer (IH-R in Table~6 of Appendix~D is this difference, grounded twin minus circular system).

As Table~2 shows, the resolution score was 0.31 standard deviations higher in grounded twins than in circular systems. The difference remained at 0.27 after adjustment for the small length difference between the systems and at 0.29 when compared with perplexity-matched controls. It appeared in both the truth wording (0.25) and the bit wording (0.37) and was significant in each model individually ($p \leq 0.0015$; Table~6). Directly using the truth direction to separate \tru{} from \fal{} correct answers (truth wording) gave an AUC of 0.69--0.85 in grounded twins. In circular systems, the AUC was only at chance level (0.44--0.51).

\begin{table}[t]
\centering\footnotesize
\setlength{\tabcolsep}{3pt}
\begin{tabular}{@{}lrr@{}}
\toprule
Prediction & Estimate [95\% CI] & $p$ \\
\midrule
P1 \inc & \PrimEst\ \PrimCI & \PrimP \\
P2 resolution score & \IHREst\ \IHRCI & \IHRP \\
\hspace{0.5em}One-step families & \IHRRoundOneEst\ \IHRRoundOneCI & $\IHRRoundOneP$ \\
\hspace{0.5em}Propagation families & \IHRRoundMultiEst\ \IHRRoundMultiCI & $\IHRRoundMultiP$ \\
P3 late-layer update & +0.11 [$-$0.50, 0.73] & $\IHUP$ \\
P4 attention rank & $-$0.94 [$-$1.35, $-$0.56] & $\IHAP$ \\
P5 resolution score & \IHRCnfEst\ \IHRCnfCI & $\IHRCnfP$ \\
\bottomrule
\end{tabular}
\caption{Results of testing P1--P5 on the 56 core families against grounded twins (P1--P4) and constraint twins (P5, truth wording), in percentage points (P1), calibration standard deviations (P2, P5), and $10^{-3}$ (P3, P4); positive values are in the predicted direction. P1 and P2 were supported, whereas P3, P4, and P5 were not (P4 in the opposite direction). The $p$-values are Holm-adjusted within the behavioral test family (P1) and within the internal test family (P2--P4), and unadjusted for P5 and the indented rows (§6.4).}
\label{tab:predictions}
\end{table}

\paragraph{Constraint twins (P5).}
P5, the prediction that constraint twins would show the same pattern as grounded twins, was not supported. As Table~2 shows, the resolution score of constraint twins did not differ significantly from that of circular systems (Table~6 gives the other two measures), and the truth direction discriminated between their \tru{} and \fal{} correct answers at approximately chance level (AUC 0.52--0.60). Even so, constraint twins drew far fewer \inc{} verdicts (§5.2; §7).

\subsection{Late-layer update and attention rank (P3,~P4)}
The predictions of continued state revision in late layers for unresolved systems (P3) and broader attention spread (P4) were not supported. For P3, the late-layer update at the question's last token was not larger in circular systems than in grounded twins (difference in relative update magnitude $+0.0001$, $p=0.733$). Length adjustment did not change this result, and no per-model test was significant. Behind the pooled result, the two wordings diverged: the difference was $+0.0015$ in the truth wording ($+0.0018$ against perplexity-matched controls, $p=9\times 10^{-5}$) and $-0.0013$ in the bit wording. When we recomputed the update without the final step (the last layer and final normalization), it was instead smaller in circular systems than in grounded twins ($-0.0012$, $p=0.0023$).

For P4, contrary to the prediction, attention rank was slightly lower in circular systems ($-0.0009$ on a 0--1 scale, $p=6\times 10^{-5}$). The direction did not change after length adjustment or in the comparison with perplexity-matched controls. Attention rank in circular systems was higher than in constraint twins ($+0.0025$, $p<10^{-5}$), but the sign of the difference from grounded twins varied across depths and models (Llama-11B was in the predicted direction).

\subsection{Within-model classification and cross-model transfer}
With many internal metrics and few prompts, a classifier can separate conditions without learning a generalizable boundary. We therefore fixed 15 input features in advance and standardized each within each model, without labels, before splitting (Appendix~C). The features are five resolution scores averaged over validated layers in five depth bins of roughly equal size, five residual updates in similarly divided bins, and five attention ranks at five depths.

The classification task is pairwise: given the feature difference between a circular system and its grounded twin, with the pair members in random order, the classifier identifies which member is the circular system. Within each model, family-grouped nested cross-validation (eight outer folds, with regularization strength selected in four inner folds), rerun on each of 1,000 family-level label permutations, gave a cross-validated AUC of 0.98--1.00 (all $p \leq 0.001$ in the permutation test).

However, circular systems and grounded twins differ in their text, and some features that best separate them within one model reverse direction across architectures. Attention rank at the earliest depth is higher for circular systems in Gemma but lower in Llama-11B and Llama-70B. When tested individually (Table~7), 13 of the 15 features differ significantly. The features with the largest absolute standardized differences $(d_z)$ for circular systems minus grounded twins are attention rank at depth 0.3 ($-2.52$), residual update in the shallowest bin ($+0.90$), and attention rank at depth 0.7 ($-0.87$).

Trained on three models and tested on the remaining model using the same 56 families, the classifier reached high AUC on Llama-70B (AUC 0.98, $p=0.036$) and Qwen (AUC 0.96, $p=0.063$). It failed to identify circular systems in Gemma (AUC 0.05, reversed direction) and Llama-11B (AUC 0.32; Table~5). Transfer remained partial both when only the top three features by importance on the training models were retained and when feature scaling was refit on training rows only (Tables~5 and~8). The classifier is therefore not a universal detector.

The resolution score difference was consistent across all four models. It is defined identically in every model, is positive and significant in each, and has the same sign in the three deepest bins. These are the only bins with validated layers in every model. An exploratory classifier using only the resolution score features is described in Appendix~C.

\subsection{One-step and propagation families}
The resolution score difference comes from one-step families, whereas the behavioral contrast is the same in both groups. In the 27 one-step families, the grounded twin's queried utterance is settled by its own constants and therefore in the first round under the evaluation procedure in §2 (Appendix~C). In the 29 propagation families, settling that value requires two to four rounds. In the one-step families, the resolution score difference is $+0.61$ standard deviations (95\% CI [0.38, 0.86], $p=3\times 10^{-5}$; Table~6). By contrast, in the propagation families, it is $+0.03$ standard deviations (95\% CI [$-$0.11, 0.18], $p=0.697$), which is not significantly different from $0$. The estimate for one-step families is larger in all four models ($+0.32$ to $+1.39$ vs.\ $+0.01$ to $+0.05$ standard deviations) and in both wordings (truth wording $+0.50$ vs.\ $+0.02$, bit wording $+0.72$ vs.\ $+0.04$).

The truth direction also separates the correct answers of grounded twins (truth wording) only weakly and inconsistently in propagation families (AUC 0.36--0.71, compared with 0.84--0.95 in one-step families). At the layer level (Figure~3b--e), the difference is positive at every validated layer of every model for one-step families (at least $+0.09$ standard deviations) and never exceeds $0.16$ standard deviations in magnitude for propagation families. However, the behavioral contrast is large in both groups (one-step 31.0 and propagation 31.0 percentage points; difference 0.0 percentage points, 95\% CI [$-$10.2, 9.8], permutation test $p=1.00$).

\section{Discussion}

\subsection{Circularity and the verdict}

Constraint twins have no truth recursion. Thus, their solutions cannot be read off through propagation; finding a solution requires search. As in circular systems, the queried value in constraint twins is not recovered along the truth direction, yet models judged constraint twins to be \inc{} much more rarely. Although accuracy was low at 18\%, the most common incorrect answer across the four models combined was \und{} rather than \inc{}. That is, because constraint twins are relatively difficult, models produce more incorrect answers, but even when wrong, they seldom classify these systems as contradictions.

Of the three item types, only circular systems have truth recursion that is unresolved at finite depth, and only they show a substantial increase in phantom paradoxes. This pattern indicates that models treat truth recursion whose values propagation cannot settle as a sign of contradiction. This heuristic is correct for the liar paradox but always wrong for satisfiable cycles. Difficulty alone does not trigger it, and it is triggered less often by self-reference whose value is settled (§5).

\paragraph{Finite-depth resolution versus convergence.}
Of the 56 core circular systems, 15 converge from every initial assignment, but none of them is grounded (§2). In a comparison not fixed in advance, each model returned slightly fewer \inc{} verdicts for these 15 systems than for the remaining 41. However, except in \mbox{Llama-11B}, which often judged even grounded twins to be \inc{}, these 15 systems still received \inc{} verdicts much more often than grounded twins (rates of \inc{} answers for converging circular systems, nonconverging circular systems, and grounded twins: Gemma 0.57/0.63/0.04, \mbox{Llama-11B} 0.53/0.56/0.44, \mbox{Llama-70B} 0.43/0.55/0.08, Qwen 0.10/0.13/0.01).

The models' verdicts therefore appear to reflect primarily whether truth recursion is resolved from the bottom up by propagation starting from unknown, rather than whether synchronous reevaluation converges from every initial assignment.

\subsection{Middle layers and the information bottleneck}
According to information bottleneck theory, a neural network compresses its input across layers into the information needed for the output (\citealp{tishby2015deep}). From this perspective, when truth recursion is resolved, we can expect the queried utterance's truth value to be recovered along the truth direction somewhere at intermediate depth.

This expectation held mainly in one-step families, whose grounded twins recovered the value at most of the validated layers. Recovery was weak and inconsistent in grounded twins of propagation families and absent in circular systems and constraint twins (§6). Similarly, models perform the first step of two-step factual questions more reliably (\citealp{yang2024latent}) and at earlier layers (\citealp{biran2024hopping}) than they perform the second. Values requiring propagation may be computed during response generation rather than while reading the prompt. We did not measure compression itself, so this account is an interpretation of the layerwise pattern rather than a tested mechanism. These measurements alone cannot identify the basis for the verdict.

\subsection{Practical implications}

\emph{Evaluation.} Benchmarks of paradox handling should include satisfiable circular systems and grounded twins. If a benchmark includes only unsatisfiable cycles, a model receives a high score by answering ``paradox'' whenever truth refers to truth. Our models answered \inc{} on 64\% of unsatisfiable circular systems but also on 43\% of satisfiable ones.

\emph{Mitigation.} Replacing circular truth statements with equivalent constraints reduced the rate of \inc{} answers by 36.4 percentage points on average (see Table~4 for results by model). However, accuracy also dropped from 32\% to 18\%, and correct \inc{} verdicts on the 36 systems with no consistent assignment fell from 64\% to 16\%. Translating self-referential rules into constraints and passing them to an external solver, as in solver-augmented pipelines (\citealp{olausson2023linc}; \citealp{pan2023logiclm}; \citealp{ye2023satlm}), may therefore be more appropriate than filtering them out.

\emph{Detection.} The truth-direction readout requires only a single projection at each validated layer, but the resolution score difference comes from one-step families (§6). Within circular systems, the resolution score for rows answered \inc{} is lower than that for other rows in Qwen and \mbox{Llama-70B}, similar in Llama-11B, and higher in Gemma. The readout therefore does not consistently predict individual errors and cannot by itself check for unresolved truth recursion. A detector would need model-specific calibration and an additional decision criterion that uses neither the correct answer nor a twin.

\section{Related Work}

\paragraph{Truth, self-reference, and groundedness.}
We defined finite-depth resolution through the least fixed point of \citet{kripke1975outline}: a self-referential sentence is grounded when evaluation starting from unknown settles its value. Kripke's theory distinguishes groundedness from paradoxicality: the truth-teller sentence is ungrounded but not paradoxical. Other work defines groundedness through semantic dependence within classical semantics (\citealp{leitgeb2005truth}). In a finite reference graph, a paradox requires a directed cycle (in infinite graphs, paradoxes arise even without cycles; \citealp{yablo1993paradox}). A cycle alone is not sufficient, however, because the liar and truth-teller sentences have the same graph (\citealp{rabern2013dangerous}; \citealp{beringer2017graph}). These theories classify sentences; our twins implement this distinction between groundedness and the existence of a consistent assignment through matched pairs.

\paragraph{Self-referential reasoning in language models.}
The puzzles in the cited knights-and-knaves benchmarks (\citealp{mondorf2024truthquest}; \citealp{xie2024memorization}) are all circular in the sense defined in §2, because every speaker's statement refers to some speaker, and they vary in difficulty (§1). Our grounded twins replace only the cycle-closing references with constants while preserving the solution, whereas our constraint twins eliminate truth recursion while preserving the logic.

ParaSuite evaluates whether models can identify and explain the contradictions in paradoxes (\citealp{chen2026parasuite}). Phantom paradoxes are the opposite error. Deductive reasoning benchmarks (\citealp{clark2020transformers}; \citealp{tafjord2021proofwriter}; \citealp{saparov2023greedy}; \citealp{han2022folio}) test reasoning from premises to conclusions, and LogicNLI added a ``Paradox'' label for cases in which both a statement and its negation are derived from the premises (\citealp{tian2021logicnli}). However, none of them tests circular truth dependence. In constraint satisfaction benchmarks without self-reference, accuracy declines as the search space grows (\citealp{lin2025zebralogic}), and language models tend to report that a solution exists (\citealp{wei2025satbench}; \citealp{zhang2026matchedpair}). The latter tendency also holds for our constraint twins (§7.1; §7.3). Metalinguistic self-reference to phrases within the same sentence is also difficult for language models to process (\citealp{thrush2024strange}), but it differs in that it lacks truth recursion.

\paragraph{Probability and surface form.}
Language-model accuracy depends heavily on the probabilities of the task and target output and, under some conditions, is also affected by input probability (\citealp{mccoy2024embers}). Models tend to perform better on prompts to which they assign lower perplexity (\citealp{gonen2023demystifying}), and few-shot numerical reasoning is more accurate on items containing terms that occur frequently in the pretraining corpus (\citealp{razeghi2022impact}). We therefore matched controls on each model's likelihood before generating responses to test whether the effect arises from the rarity of self-referential text.

\paragraph{Truth directions and probing.}
Classifiers trained on hidden-layer activations distinguish true from false statements (\citealp{azaria2023internal}), often through a linear direction (\citealp{burns2023discovering}; \citealp{li2023inference}; \citealp{marks2024geometry}; \citealp{burger2024truth}). Directions fitted to simple declarative sentences can transfer to logically transformed statements and to question answering, and transfer to negated statements is more consistent in more capable models (\citealp{bao2025geometry}). However, transfer across tasks is uneven (\citealp{azizian2025geometries}), and results also vary with the layer probed (\citealp{poulis2026limits}).

Because probing results are easy to overinterpret (\citealp{hewitt2019control}; \citealp{belinkov2022probing}), we use only truth directions that have been validated layer by layer (§6) and compare circular systems with grounded twins that share the same correct answer.

\section{Conclusion}

We analyzed phantom paradoxes, in which satisfiable circular systems are mistaken for contradictions, across four language models. All four models reported contradictions more often on circular systems: on average, 31.0 percentage points more often than on grounded twins with the same solution and 36.4 percentage points more often than on constraint twins. This tendency persisted after controlling for input perplexity and wording.

In the internal analysis, the queried truth value was not recovered in circular systems, but changes in truth-value recovery did not align with changes in contradiction verdicts. This shows that phantom paradoxes cannot be explained solely by a failure to linearly read out correct-answer information. Together with the behavioral contrasts, it also shows that models treat truth recursion that propagation cannot settle as a signal of contradiction.

These findings raise the question of when language models treat systems whose truth values are difficult to settle as unsatisfiable. They also show that evaluations of paradox handling need to include satisfiable circular systems. Further research is needed to clarify in detail how truth-dependence structures affect final contradiction verdicts.

\section*{Limitations}

We tested four open models from three families with English prompts, using greedy decoding and a single prompt format. Accuracy and model rankings can vary substantially with prompt formatting and paraphrasing (\citealp{sclar2024quantifying}; \citealp{mizrahi2024state}). Reasoning models trained on knights-and-knaves puzzles (\citealp{xie2025logicrl}) or closed models may behave differently.

Qwen is the most accurate model on all three item types (Table~3) and one of the two models with the smallest difference between circular systems and grounded twins. It is also the only tested model that applies QK normalization in self-attention (\citealp{qwen3}; \citealp{qwen3vl}). The effect size does not vary monotonically with depth (32--80 layers). A sample of four models cannot determine whether QK normalization or depth changes the effect, so we do not attribute cross-model differences to architecture.

Our truth direction is a linear direction fitted to arithmetic propositions, and such directions transfer unevenly across tasks (\citealp{azizian2025geometries}; \citealp{poulis2026limits}). Failure to read out a value along this direction does not rule out its representation in another form. Our internal analysis is correlational and does not intervene on the residual stream. The analysis reads only the prompt state before reasoning is generated, excluding values computed later during generation, as may occur in propagation families (§6.4; §7.2).

Grounded twins remove cycles while making the queried utterance's value reachable through local propagation alone, so the effects of these changes are hard to separate. Constraint twins and perplexity-matched controls address the most obvious alternative explanations (logic and rarity), but cannot address every difference arising from surface processing. In the main dataset, circularity and unresolved truth recursion always co-occur. The only evidence that verdicts vary with groundedness when self-reference is held fixed comes from the small natural-language dataset (§5).

\bibliography{references}

\begin{thebibliography}{53}
\expandafter\ifx\csname natexlab\endcsname\relax\def\natexlab#1{#1}\fi

\bibitem[{Azaria and Mitchell(2023)}]{azaria2023internal}
Amos Azaria and Tom Mitchell. 2023.
\newblock \href {https://doi.org/10.18653/v1/2023.findings-emnlp.68} {The
  internal state of an {LLM} knows when it's lying}.
\newblock In \emph{Findings of the Association for Computational Linguistics:
  EMNLP 2023}, pages 967--976, Singapore. Association for Computational
  Linguistics.

\bibitem[{Azizian et~al.(2025)Azizian, Kirchhof, Ndiaye, Bethune, Klein, Ablin,
  and Cuturi}]{azizian2025geometries}
Waiss Azizian, Michael Kirchhof, Eugene Ndiaye, Louis Bethune, Michal Klein,
  Pierre Ablin, and Marco Cuturi. 2025.
\newblock \href {http://arxiv.org/abs/2506.08572} {The geometries of truth are
  orthogonal across tasks}.
\newblock arXiv preprint arXiv:2506.08572v2.
\newblock Presented at the ICML 2025 Workshop on Reliable and Responsible
  Foundation Models.

\bibitem[{Bai et~al.(2025)Bai, Cai, Chen, Chen, Chen, Cheng, Deng, Ding, Gao,
  Ge, Ge, Guo, Huang, Huang, Huang, Hui, Jiang, Li, Li, Li, Li, Lin, Lin, Liu,
  Liu, Liu, Liu, Liu, Liu, Lu, Luo, Lv, Men, Meng, Ren, Ren, Song, Sun, Tang,
  Tu, Wan, Wang, Wang, Wang, Wang, Xie, Xu, Xu, Xu, Yang, Yang, Yang, Yang, Yu,
  Zhang, Zhang, Zhang, Zheng, Zhong, Zhou, Zhou, Zhou, Zhu, and Zhu}]{qwen3vl}
Shuai Bai, Yuxuan Cai, Ruizhe Chen, Keqin Chen, Xionghui Chen, Zesen Cheng,
  Lianghao Deng, Wei Ding, Chang Gao, Chunjiang Ge, Wenbin Ge, Zhifang Guo,
  Qidong Huang, Jie Huang, Fei Huang, Binyuan Hui, Shutong Jiang, Zhaohai Li,
  Mingsheng Li, Mei Li, Kaixin Li, Zicheng Lin, Junyang Lin, Xuejing Liu,
  Jiawei Liu, Chenglong Liu, Yang Liu, Dayiheng Liu, Shixuan Liu, Dunjie Lu,
  Ruilin Luo, Chenxu Lv, Rui Men, Lingchen Meng, Xuancheng Ren, Xingzhang Ren,
  Sibo Song, Yuchong Sun, Jun Tang, Jianhong Tu, Jianqiang Wan, Peng Wang,
  Pengfei Wang, Qiuyue Wang, Yuxuan Wang, Tianbao Xie, Yiheng Xu, Haiyang Xu,
  Jin Xu, Zhibo Yang, Mingkun Yang, Jianxin Yang, An~Yang, Bowen Yu, Fei Zhang,
  Hang Zhang, Xi~Zhang, Bo~Zheng, Humen Zhong, Jingren Zhou, Fan Zhou, Jing
  Zhou, Yuanzhi Zhu, and Ke~Zhu. 2025.
\newblock \href {http://arxiv.org/abs/2511.21631} {{Qwen3-VL} technical
  report}.
\newblock arXiv preprint arXiv:2511.21631v2.

\bibitem[{Bao et~al.(2025)Bao, Zhang, Du, Zhao, Feng, Peng, and
  Yin}]{bao2025geometry}
Yuntai Bao, Xuhong Zhang, Tianyu Du, Xinkui Zhao, Zhengwen Feng, Hao Peng, and
  Jianwei Yin. 2025.
\newblock \href {https://doi.org/10.18653/v1/2025.findings-acl.38} {Probing the
  geometry of truth: {Consistency} and generalization of truth directions in
  {LLM}s across logical transformations and question answering tasks}.
\newblock In \emph{Findings of the Association for Computational Linguistics:
  ACL 2025}, pages 682--700, Vienna, Austria. Association for Computational
  Linguistics.

\bibitem[{Belinkov(2022)}]{belinkov2022probing}
Yonatan Belinkov. 2022.
\newblock \href {https://doi.org/10.1162/coli_a_00422} {Probing classifiers:
  {Promises}, shortcomings, and advances}.
\newblock \emph{Computational Linguistics}, 48(1):207--219.

\bibitem[{Berg-Kirkpatrick et~al.(2012)Berg-Kirkpatrick, Burkett, and
  Klein}]{bergkirkpatrick2012empirical}
Taylor Berg-Kirkpatrick, David Burkett, and Dan Klein. 2012.
\newblock An empirical investigation of statistical significance in {NLP}.
\newblock In \emph{Proceedings of the 2012 Joint Conference on Empirical
  Methods in Natural Language Processing and Computational Natural Language
  Learning}, pages 995--1005, Jeju Island, Korea. Association for Computational
  Linguistics.

\bibitem[{Beringer and Schindler(2017)}]{beringer2017graph}
Timo Beringer and Thomas Schindler. 2017.
\newblock \href {https://doi.org/10.1017/bsl.2017.37} {A graph-theoretic
  analysis of the semantic paradoxes}.
\newblock \emph{The Bulletin of Symbolic Logic}, 23(4):442--492.

\bibitem[{Biran et~al.(2024)Biran, Gottesman, Yang, Geva, and
  Globerson}]{biran2024hopping}
Eden Biran, Daniela Gottesman, Sohee Yang, Mor Geva, and Amir Globerson. 2024.
\newblock \href {https://doi.org/10.18653/v1/2024.emnlp-main.781} {Hopping too
  late: Exploring the limitations of large language models on multi-hop
  queries}.
\newblock In \emph{Proceedings of the 2024 Conference on Empirical Methods in
  Natural Language Processing}, pages 14113--14130, Miami, Florida, USA.
  Association for Computational Linguistics.

\bibitem[{B{\"u}rger et~al.(2024)B{\"u}rger, Hamprecht, and
  Nadler}]{burger2024truth}
Lennart B{\"u}rger, Fred~A. Hamprecht, and Boaz Nadler. 2024.
\newblock \href {https://doi.org/10.52202/079017-4392} {Truth is universal:
  {Robust} detection of lies in {LLM}s}.
\newblock In \emph{Advances in Neural Information Processing Systems},
  volume~37, pages 138393--138431. Curran Associates, Inc.

\bibitem[{Burns et~al.(2023)Burns, Ye, Klein, and
  Steinhardt}]{burns2023discovering}
Collin Burns, Haotian Ye, Dan Klein, and Jacob Steinhardt. 2023.
\newblock Discovering latent knowledge in language models without supervision.
\newblock In \emph{The Eleventh International Conference on Learning
  Representations}.

\bibitem[{Chen et~al.(2026)Chen, Zhang, Ye, Wang, Liu, and
  Chen}]{chen2026parasuite}
Bin Chen, Yu~Zhang, Hongfei Ye, Huiyang Wang, Wenxi Liu, and Hongyang Chen.
  2026.
\newblock \href {https://doi.org/10.18653/v1/2026.acl-long.2047} {{ParaSuite}:
  {Boosting} {LLM} reasoning via paradox resolution}.
\newblock In \emph{Proceedings of the 64th Annual Meeting of the Association
  for Computational Linguistics (Volume 1: Long Papers)}, pages 44246--44256,
  San Diego, California, United States. Association for Computational
  Linguistics.

\bibitem[{Clark et~al.(2020)Clark, Tafjord, and
  Richardson}]{clark2020transformers}
Peter Clark, Oyvind Tafjord, and Kyle Richardson. 2020.
\newblock \href {https://doi.org/10.24963/ijcai.2020/537} {Transformers as soft
  reasoners over language}.
\newblock In \emph{Proceedings of the Twenty-Ninth International Joint
  Conference on Artificial Intelligence}, pages 3882--3890. International Joint
  Conferences on Artificial Intelligence Organization.

\bibitem[{Dror et~al.(2017)Dror, Baumer, Bogomolov, and
  Reichart}]{dror2017replicability}
Rotem Dror, Gili Baumer, Marina Bogomolov, and Roi Reichart. 2017.
\newblock \href {https://doi.org/10.1162/tacl_a_00074} {Replicability analysis
  for natural language processing: {Testing} significance with multiple
  datasets}.
\newblock \emph{Transactions of the Association for Computational Linguistics},
  5:471--486.

\bibitem[{Dror et~al.(2018)Dror, Baumer, Shlomov, and
  Reichart}]{dror2018hitchhiker}
Rotem Dror, Gili Baumer, Segev Shlomov, and Roi Reichart. 2018.
\newblock \href {https://doi.org/10.18653/v1/P18-1128} {The hitchhiker's guide
  to testing statistical significance in natural language processing}.
\newblock In \emph{Proceedings of the 56th Annual Meeting of the Association
  for Computational Linguistics (Volume 1: Long Papers)}, pages 1383--1392,
  Melbourne, Australia. Association for Computational Linguistics.

\bibitem[{{Gemma Team} et~al.(2024){Gemma Team}, Riviere, Pathak, Sessa,
  Hardin, Bhupatiraju, Hussenot, Mesnard, Shahriari, Ram{\'e}, Ferret, Liu,
  Tafti, Friesen, Casbon, Ramos, Kumar, Lan, Jerome, Tsitsulin, Vieillard,
  Stanczyk, Girgin, Momchev, Hoffman, Thakoor, Grill, Neyshabur, Bachem,
  Walton, Severyn, Parrish, Ahmad, Hutchison, Abdagic, Carl, Shen, Brock,
  Coenen, Laforge, Paterson, Bastian, Piot, Wu, Royal, Chen, Kumar, Perry,
  Welty, Choquette-Choo, Sinopalnikov, Weinberger, Vijaykumar, Rogozi{\'n}ska,
  Herbison, Bandy, Wang, Noland, Moreira, Senter, Eltyshev, Visin, Rasskin,
  Wei, Cameron, Martins, Hashemi, Klimczak-Pluci{\'n}ska, Batra, Dhand,
  Nardini, Mein, Zhou, Svensson, Stanway, Chan, Zhou, Carrasqueira, Iljazi,
  Becker, Fernandez, van Amersfoort, Gordon, Lipschultz, Newlan, Ji, Mohamed,
  Badola, Black, Millican, McDonell, Nguyen, Sodhia, Greene, Sjoesund, Usui,
  Sifre, Heuermann, Lago, McNealus, Soares, Kilpatrick, Dixon, Martins, Reid,
  Singh, Iverson, G{\"o}rner, Velloso, Wirth, Davidow, Miller, Rahtz, Watson,
  Risdal, Kazemi, Moynihan, Zhang, Kahng, Park, Rahman, Khatwani, Dao,
  Bardoliwalla, Devanathan, Dumai, Chauhan, Wahltinez, Botarda, Barnes, Barham,
  Michel, Jin, Georgiev, Culliton, Kuppala, Comanescu, Merhej, Jana, Rokni,
  Agarwal, Mullins, Saadat, Carthy, Cogan, Perrin, Arnold, Krause, Dai, Garg,
  Sheth, Ronstrom, Chan, Jordan, Yu, Eccles, Hennigan, Kocisky, Doshi, Jain,
  Yadav, Meshram, Dharmadhikari, Barkley, Wei, Ye, Han, Kwon, Xu, Shen, Gong,
  Wei, Cotruta, Kirk, Rao, Giang, Peran, Warkentin, Collins, Barral,
  Ghahramani, Hadsell, Sculley, Banks, Dragan, Petrov, Vinyals, Dean, Hassabis,
  Kavukcuoglu, Farabet, Buchatskaya, Borgeaud, Fiedel, Joulin, Kenealy,
  Dadashi, and Andreev}]{gemma2}
{Gemma Team}, Morgane Riviere, Shreya Pathak, Pier~Giuseppe Sessa, Cassidy
  Hardin, Surya Bhupatiraju, L{\'e}onard Hussenot, Thomas Mesnard, Bobak
  Shahriari, Alexandre Ram{\'e}, Johan Ferret, Peter Liu, Pouya Tafti, Abe
  Friesen, Michelle Casbon, Sabela Ramos, Ravin Kumar, Charline~Le Lan, Sammy
  Jerome, Anton Tsitsulin, Nino Vieillard, Piotr Stanczyk, Sertan Girgin,
  Nikola Momchev, Matt Hoffman, Shantanu Thakoor, Jean-Bastien Grill, Behnam
  Neyshabur, Olivier Bachem, Alanna Walton, Aliaksei Severyn, Alicia Parrish,
  Aliya Ahmad, Allen Hutchison, Alvin Abdagic, Amanda Carl, Amy Shen, Andy
  Brock, Andy Coenen, Anthony Laforge, Antonia Paterson, Ben Bastian, Bilal
  Piot, Bo~Wu, Brandon Royal, Charlie Chen, Chintu Kumar, Chris Perry, Chris
  Welty, Christopher~A. Choquette-Choo, Danila Sinopalnikov, David Weinberger,
  Dimple Vijaykumar, Dominika Rogozi{\'n}ska, Dustin Herbison, Elisa Bandy,
  Emma Wang, Eric Noland, Erica Moreira, Evan Senter, Evgenii Eltyshev,
  Francesco Visin, Gabriel Rasskin, Gary Wei, Glenn Cameron, Gus Martins, Hadi
  Hashemi, Hanna Klimczak-Pluci{\'n}ska, Harleen Batra, Harsh Dhand, Ivan
  Nardini, Jacinda Mein, Jack Zhou, James Svensson, Jeff Stanway, Jetha Chan,
  Jin~Peng Zhou, Joana Carrasqueira, Joana Iljazi, Jocelyn Becker, Joe
  Fernandez, Joost van Amersfoort, Josh Gordon, Josh Lipschultz, Josh Newlan,
  {Ju-yeong} Ji, Kareem Mohamed, Kartikeya Badola, Kat Black, Katie Millican,
  Keelin McDonell, Kelvin Nguyen, Kiranbir Sodhia, Kish Greene, Lars~Lowe
  Sjoesund, Lauren Usui, Laurent Sifre, Lena Heuermann, Leticia Lago, Lilly
  McNealus, Livio~Baldini Soares, Logan Kilpatrick, Lucas Dixon, Luciano
  Martins, Machel Reid, Manvinder Singh, Mark Iverson, Martin G{\"o}rner, Mat
  Velloso, Mateo Wirth, Matt Davidow, Matt Miller, Matthew Rahtz, Matthew
  Watson, Meg Risdal, Mehran Kazemi, Michael Moynihan, Ming Zhang, Minsuk
  Kahng, Minwoo Park, Mofi Rahman, Mohit Khatwani, Natalie Dao, Nenshad
  Bardoliwalla, Nesh Devanathan, Neta Dumai, Nilay Chauhan, Oscar Wahltinez,
  Pankil Botarda, Parker Barnes, Paul Barham, Paul Michel, Pengchong Jin, Petko
  Georgiev, Phil Culliton, Pradeep Kuppala, Ramona Comanescu, Ramona Merhej,
  Reena Jana, Reza~Ardeshir Rokni, Rishabh Agarwal, Ryan Mullins, Samaneh
  Saadat, Sara~Mc Carthy, Sarah Cogan, Sarah Perrin, S{\'e}bastien M.~R.
  Arnold, Sebastian Krause, Shengyang Dai, Shruti Garg, Shruti Sheth, Sue
  Ronstrom, Susan Chan, Timothy Jordan, Ting Yu, Tom Eccles, Tom Hennigan,
  Tomas Kocisky, Tulsee Doshi, Vihan Jain, Vikas Yadav, Vilobh Meshram, Vishal
  Dharmadhikari, Warren Barkley, Wei Wei, Wenming Ye, Woohyun Han, Woosuk Kwon,
  Xiang Xu, Zhe Shen, Zhitao Gong, Zichuan Wei, Victor Cotruta, Phoebe Kirk,
  Anand Rao, Minh Giang, Ludovic Peran, Tris Warkentin, Eli Collins, Joelle
  Barral, Zoubin Ghahramani, Raia Hadsell, D.~Sculley, Jeanine Banks, Anca
  Dragan, Slav Petrov, Oriol Vinyals, Jeff Dean, Demis Hassabis, Koray
  Kavukcuoglu, Clement Farabet, Elena Buchatskaya, Sebastian Borgeaud, Noah
  Fiedel, Armand Joulin, Kathleen Kenealy, Robert Dadashi, and Alek Andreev.
  2024.
\newblock \href {http://arxiv.org/abs/2408.00118} {{Gemma} 2: {Improving} open
  language models at a practical size}.
\newblock arXiv preprint arXiv:2408.00118v3.

\bibitem[{Gonen et~al.(2023)Gonen, Iyer, Blevins, Smith, and
  Zettlemoyer}]{gonen2023demystifying}
Hila Gonen, Srini Iyer, Terra Blevins, Noah~A. Smith, and Luke Zettlemoyer.
  2023.
\newblock \href {https://doi.org/10.18653/v1/2023.findings-emnlp.679}
  {Demystifying prompts in language models via perplexity estimation}.
\newblock In \emph{Findings of the Association for Computational Linguistics:
  EMNLP 2023}, pages 10136--10148, Singapore. Association for Computational
  Linguistics.

\bibitem[{Grattafiori et~al.(2024)Grattafiori, Dubey, Jauhri, Pandey, Kadian,
  Al-Dahle, Letman, Mathur, Schelten, Vaughan, Yang, Fan, Goyal, Hartshorn,
  Yang, Mitra, Sravankumar, Korenev, Hinsvark, Rao, Zhang, Rodriguez,
  Gregerson, Spataru, Roziere, Biron, Tang, Chern, Caucheteux, Nayak, Bi,
  Marra, McConnell, Keller, Touret, Wu, Wong, Ferrer, Nikolaidis, Allonsius,
  Song, Pintz, Livshits, Wyatt, Esiobu, Choudhary, Mahajan, Garcia-Olano,
  Perino, Hupkes, Lakomkin, AlBadawy, Lobanova, Dinan, Smith, Radenovic,
  Guzm{\'a}n, Zhang, Synnaeve, Lee, Anderson, Thattai, Nail, Mialon, Pang,
  Cucurell, Nguyen, Korevaar, Xu, Touvron, Zarov, Ibarra, Kloumann, Misra,
  Evtimov, Zhang, Copet, Lee, Geffert, Vranes, Park, Mahadeokar, Shah, van~der
  Linde, Billock, Hong, Lee, Fu, Chi, Huang, Liu, Wang, Yu, Bitton, Spisak,
  Park, Rocca, Johnstun, Saxe, Jia, Alwala, Prasad, Upasani, Plawiak, Li,
  Heafield, Stone, El-Arini, Iyer, Malik, Chiu, Bhalla, Lakhotia,
  Rantala-Yeary, van~der Maaten, Chen, Tan, Jenkins, Martin, Madaan, Malo,
  Blecher, Landzaat, de~Oliveira, Muzzi, Pasupuleti, Singh, Paluri, Kardas,
  Tsimpoukelli, Oldham, Rita, Pavlova, Kambadur, Lewis, Si, Singh, Hassan,
  Goyal, Torabi, Bashlykov, Bogoychev, Chatterji, Zhang, Duchenne, {\c C}elebi,
  Alrassy, Zhang, Li, Vasic, Weng, Bhargava, Dubal, Krishnan, Koura, Xu, He,
  Dong, Srinivasan, Ganapathy, Calderer, Cabral, Stojnic, Raileanu, Maheswari,
  Girdhar, Patel, Sauvestre, Polidoro, Sumbaly, Taylor, Silva, Hou, Wang,
  Hosseini, Chennabasappa, Singh, Bell, Kim, Edunov, Nie, Narang, Raparthy,
  Shen, Wan, Bhosale, Zhang, Vandenhende, Batra, Whitman, Sootla, Collot,
  Gururangan, Borodinsky, Herman, Fowler, Sheasha, Georgiou, Scialom,
  Speckbacher, Mihaylov, Xiao, Karn, Goswami, Gupta, Ramanathan, Kerkez,
  Gonguet, Do, Vogeti, Albiero, Petrovic, Chu, Xiong, Fu, Meers, Martinet,
  Wang, Wang, Tan, Xia, Xie, Jia, Wang, Goldschlag, Gaur, Babaei, Wen, Song,
  Zhang, Li, Mao, Coudert, Yan, Chen, Papakipos, Singh, Srivastava, Jain,
  Kelsey, Shajnfeld, Gangidi, Victoria, Goldstand, Menon, Sharma, Boesenberg,
  Baevski, Feinstein, Kallet, Sangani, Teo, Yunus, Lupu, Alvarado, Caples, Gu,
  Ho, Poulton, Ryan, Ramchandani, Dong, Franco, Goyal, Saraf, Chowdhury,
  Gabriel, Bharambe, Eisenman, Yazdan, James, Maurer, Leonhardi, Huang, Loyd,
  Paola, Paranjape, Liu, Wu, Ni, Hancock, Wasti, Spence, Stojkovic, Gamido,
  Montalvo, Parker, Burton, Mejia, Liu, Wang, Kim, Zhou, Hu, Chu, Cai, Tindal,
  Feichtenhofer, Gao, Civin, Beaty, Kreymer, Li, Adkins, Xu, Testuggine, David,
  Parikh, Liskovich, Foss, Wang, Le, Holland, Dowling, Jamil, Montgomery,
  Presani, Hahn, Wood, Le, Brinkman, Arcaute, Dunbar, Smothers, Sun, Kreuk,
  Tian, Kokkinos, Ozgenel, Caggioni, Kanayet, Seide, Florez, Schwarz, Badeer,
  Swee, Halpern, Herman, Sizov, Zhang, Lakshminarayanan, Inan, Shojanazeri,
  Zou, Wang, Zha, Habeeb, Rudolph, Suk, Aspegren, Goldman, Zhan, Damlaj,
  Molybog, Tufanov, Leontiadis, Veliche, Gat, Weissman, Geboski, Kohli, Lam,
  Asher, Gaya, Marcus, Tang, Chan, Zhen, Reizenstein, Teboul, Zhong, Jin, Yang,
  Cummings, Carvill, Shepard, McPhie, Torres, Ginsburg, Wang, Wu, U, Saxena,
  Khandelwal, Zand, Matosich, Veeraraghavan, Michelena, Li, Jagadeesh, Huang,
  Chawla, Huang, Chen, Garg, A, Silva, Bell, Zhang, Guo, Yu, Moshkovich,
  Wehrstedt, Khabsa, Avalani, Bhatt, Mankus, Hasson, Lennie, Reso, Groshev,
  Naumov, Lathi, Keneally, Liu, Seltzer, Valko, Restrepo, Patel, Vyatskov,
  Samvelyan, Clark, Macey, Wang, Hermoso, Metanat, Rastegari, Bansal,
  Santhanam, Parks, White, Bawa, Singhal, Egebo, Usunier, Mehta, Laptev, Dong,
  Cheng, Chernoguz, Hart, Salpekar, Kalinli, Kent, Parekh, Saab, Balaji,
  Rittner, Bontrager, Roux, Dollar, Zvyagina, Ratanchandani, Yuvraj, Liang,
  Alao, Rodriguez, Ayub, Murthy, Nayani, Mitra, Parthasarathy, Li, Hogan,
  Battey, Wang, Howes, Rinott, Mehta, Siby, Bondu, Datta, Chugh, Hunt, Dhillon,
  Sidorov, Pan, Mahajan, Verma, Yamamoto, Ramaswamy, Lindsay, Lindsay, Feng,
  Lin, Zha, Patil, Shankar, Zhang, Zhang, Wang, Agarwal, Sajuyigbe, Chintala,
  Max, Chen, Kehoe, Satterfield, Govindaprasad, Gupta, Deng, Cho, Virk,
  Subramanian, Choudhury, Goldman, Remez, Glaser, Best, Koehler, Robinson, Li,
  Zhang, Matthews, Chou, Shaked, Vontimitta, Ajayi, Montanez, Mohan, Kumar,
  Mangla, Ionescu, Poenaru, Mihailescu, Ivanov, Li, Wang, Jiang, Bouaziz,
  Constable, Tang, Wu, Wang, Wu, Gao, Kleinman, Chen, Hu, Jia, Qi, Li, Zhang,
  Zhang, Adi, Nam, Wang, Zhao, Hao, Qian, Li, He, Rait, DeVito, Rosnbrick, Wen,
  Yang, Zhao, and Ma}]{llama3}
Aaron Grattafiori, Abhimanyu Dubey, Abhinav Jauhri, Abhinav Pandey, Abhishek
  Kadian, Ahmad Al-Dahle, Aiesha Letman, Akhil Mathur, Alan Schelten, Alex
  Vaughan, Amy Yang, Angela Fan, Anirudh Goyal, Anthony Hartshorn, Aobo Yang,
  Archi Mitra, Archie Sravankumar, Artem Korenev, Arthur Hinsvark, Arun Rao,
  Aston Zhang, Aurelien Rodriguez, Austen Gregerson, Ava Spataru, Baptiste
  Roziere, Bethany Biron, Binh Tang, Bobbie Chern, Charlotte Caucheteux, Chaya
  Nayak, Chloe Bi, Chris Marra, Chris McConnell, Christian Keller, Christophe
  Touret, Chunyang Wu, Corinne Wong, Cristian~Canton Ferrer, Cyrus Nikolaidis,
  Damien Allonsius, Daniel Song, Danielle Pintz, Danny Livshits, Danny Wyatt,
  David Esiobu, Dhruv Choudhary, Dhruv Mahajan, Diego Garcia-Olano, Diego
  Perino, Dieuwke Hupkes, Egor Lakomkin, Ehab AlBadawy, Elina Lobanova, Emily
  Dinan, Eric~Michael Smith, Filip Radenovic, Francisco Guzm{\'a}n, Frank
  Zhang, Gabriel Synnaeve, Gabrielle Lee, Georgia~Lewis Anderson, Govind
  Thattai, Graeme Nail, Gregoire Mialon, Guan Pang, Guillem Cucurell, Hailey
  Nguyen, Hannah Korevaar, Hu~Xu, Hugo Touvron, Iliyan Zarov, Imanol~Arrieta
  Ibarra, Isabel Kloumann, Ishan Misra, Ivan Evtimov, Jack Zhang, Jade Copet,
  Jaewon Lee, Jan Geffert, Jana Vranes, Jason Park, Jay Mahadeokar, Jeet Shah,
  Jelmer van~der Linde, Jennifer Billock, Jenny Hong, Jenya Lee, Jeremy Fu,
  Jianfeng Chi, Jianyu Huang, Jiawen Liu, Jie Wang, Jiecao Yu, Joanna Bitton,
  Joe Spisak, Jongsoo Park, Joseph Rocca, Joshua Johnstun, Joshua Saxe, Junteng
  Jia, Kalyan~Vasuden Alwala, Karthik Prasad, Kartikeya Upasani, Kate Plawiak,
  Ke~Li, Kenneth Heafield, Kevin Stone, Khalid El-Arini, Krithika Iyer, Kshitiz
  Malik, Kuenley Chiu, Kunal Bhalla, Kushal Lakhotia, Lauren Rantala-Yeary,
  Laurens van~der Maaten, Lawrence Chen, Liang Tan, Liz Jenkins, Louis Martin,
  Lovish Madaan, Lubo Malo, Lukas Blecher, Lukas Landzaat, Luke de~Oliveira,
  Madeline Muzzi, Mahesh Pasupuleti, Mannat Singh, Manohar Paluri, Marcin
  Kardas, Maria Tsimpoukelli, Mathew Oldham, Mathieu Rita, Maya Pavlova,
  Melanie Kambadur, Mike Lewis, Min Si, Mitesh~Kumar Singh, Mona Hassan, Naman
  Goyal, Narjes Torabi, Nikolay Bashlykov, Nikolay Bogoychev, Niladri
  Chatterji, Ning Zhang, Olivier Duchenne, Onur {\c C}elebi, Patrick Alrassy,
  Pengchuan Zhang, Pengwei Li, Petar Vasic, Peter Weng, Prajjwal Bhargava,
  Pratik Dubal, Praveen Krishnan, Punit~Singh Koura, Puxin Xu, Qing He,
  Qingxiao Dong, Ragavan Srinivasan, Raj Ganapathy, Ramon Calderer,
  Ricardo~Silveira Cabral, Robert Stojnic, Roberta Raileanu, Rohan Maheswari,
  Rohit Girdhar, Rohit Patel, Romain Sauvestre, Ronnie Polidoro, Roshan
  Sumbaly, Ross Taylor, Ruan Silva, Rui Hou, Rui Wang, Saghar Hosseini, Sahana
  Chennabasappa, Sanjay Singh, Sean Bell, Seohyun~Sonia Kim, Sergey Edunov,
  Shaoliang Nie, Sharan Narang, Sharath Raparthy, Sheng Shen, Shengye Wan,
  Shruti Bhosale, Shun Zhang, Simon Vandenhende, Soumya Batra, Spencer Whitman,
  Sten Sootla, Stephane Collot, Suchin Gururangan, Sydney Borodinsky, Tamar
  Herman, Tara Fowler, Tarek Sheasha, Thomas Georgiou, Thomas Scialom, Tobias
  Speckbacher, Todor Mihaylov, Tong Xiao, Ujjwal Karn, Vedanuj Goswami, Vibhor
  Gupta, Vignesh Ramanathan, Viktor Kerkez, Vincent Gonguet, Virginie Do, Vish
  Vogeti, V{\'\i}tor Albiero, Vladan Petrovic, Weiwei Chu, Wenhan Xiong, Wenyin
  Fu, Whitney Meers, Xavier Martinet, Xiaodong Wang, Xiaofang Wang,
  Xiaoqing~Ellen Tan, Xide Xia, Xinfeng Xie, Xuchao Jia, Xuewei Wang, Yaelle
  Goldschlag, Yashesh Gaur, Yasmine Babaei, Yi~Wen, Yiwen Song, Yuchen Zhang,
  Yue Li, Yuning Mao, Zacharie~Delpierre Coudert, Zheng Yan, Zhengxing Chen,
  Zoe Papakipos, Aaditya Singh, Aayushi Srivastava, Abha Jain, Adam Kelsey,
  Adam Shajnfeld, Adithya Gangidi, Adolfo Victoria, Ahuva Goldstand, Ajay
  Menon, Ajay Sharma, Alex Boesenberg, Alexei Baevski, Allie Feinstein, Amanda
  Kallet, Amit Sangani, Amos Teo, Anam Yunus, Andrei Lupu, Andres Alvarado,
  Andrew Caples, Andrew Gu, Andrew Ho, Andrew Poulton, Andrew Ryan, Ankit
  Ramchandani, Annie Dong, Annie Franco, Anuj Goyal, Aparajita Saraf,
  Arkabandhu Chowdhury, Ashley Gabriel, Ashwin Bharambe, Assaf Eisenman, Azadeh
  Yazdan, Beau James, Ben Maurer, Benjamin Leonhardi, Bernie Huang, Beth Loyd,
  Beto~De Paola, Bhargavi Paranjape, Bing Liu, Bo~Wu, Boyu Ni, Braden Hancock,
  Bram Wasti, Brandon Spence, Brani Stojkovic, Brian Gamido, Britt Montalvo,
  Carl Parker, Carly Burton, Catalina Mejia, Ce~Liu, Changhan Wang, Changkyu
  Kim, Chao Zhou, Chester Hu, Ching-Hsiang Chu, Chris Cai, Chris Tindal,
  Christoph Feichtenhofer, Cynthia Gao, Damon Civin, Dana Beaty, Daniel
  Kreymer, Daniel Li, David Adkins, David Xu, Davide Testuggine, Delia David,
  Devi Parikh, Diana Liskovich, Didem Foss, Dingkang Wang, Duc Le, Dustin
  Holland, Edward Dowling, Eissa Jamil, Elaine Montgomery, Eleonora Presani,
  Emily Hahn, Emily Wood, Eric-Tuan Le, Erik Brinkman, Esteban Arcaute, Evan
  Dunbar, Evan Smothers, Fei Sun, Felix Kreuk, Feng Tian, Filippos Kokkinos,
  Firat Ozgenel, Francesco Caggioni, Frank Kanayet, Frank Seide,
  Gabriela~Medina Florez, Gabriella Schwarz, Gada Badeer, Georgia Swee, Gil
  Halpern, Grant Herman, Grigory Sizov, Guangyi Zhang, Guna Lakshminarayanan,
  Hakan Inan, Hamid Shojanazeri, Han Zou, Hannah Wang, Hanwen Zha, Haroun
  Habeeb, Harrison Rudolph, Helen Suk, Henry Aspegren, Hunter Goldman, Hongyuan
  Zhan, Ibrahim Damlaj, Igor Molybog, Igor Tufanov, Ilias Leontiadis,
  Irina-Elena Veliche, Itai Gat, Jake Weissman, James Geboski, James Kohli,
  Janice Lam, Japhet Asher, Jean-Baptiste Gaya, Jeff Marcus, Jeff Tang,
  Jennifer Chan, Jenny Zhen, Jeremy Reizenstein, Jeremy Teboul, Jessica Zhong,
  Jian Jin, Jingyi Yang, Joe Cummings, Jon Carvill, Jon Shepard, Jonathan
  McPhie, Jonathan Torres, Josh Ginsburg, Junjie Wang, Kai Wu, Kam~Hou U, Karan
  Saxena, Kartikay Khandelwal, Katayoun Zand, Kathy Matosich, Kaushik
  Veeraraghavan, Kelly Michelena, Keqian Li, Kiran Jagadeesh, Kun Huang, Kunal
  Chawla, Kyle Huang, Lailin Chen, Lakshya Garg, Lavender A, Leandro Silva, Lee
  Bell, Lei Zhang, Liangpeng Guo, Licheng Yu, Liron Moshkovich, Luca Wehrstedt,
  Madian Khabsa, Manav Avalani, Manish Bhatt, Martynas Mankus, Matan Hasson,
  Matthew Lennie, Matthias Reso, Maxim Groshev, Maxim Naumov, Maya Lathi,
  Meghan Keneally, Miao Liu, Michael~L. Seltzer, Michal Valko, Michelle
  Restrepo, Mihir Patel, Mik Vyatskov, Mikayel Samvelyan, Mike Clark, Mike
  Macey, Mike Wang, Miquel~Jubert Hermoso, Mo~Metanat, Mohammad Rastegari,
  Munish Bansal, Nandhini Santhanam, Natascha Parks, Natasha White, Navyata
  Bawa, Nayan Singhal, Nick Egebo, Nicolas Usunier, Nikhil Mehta,
  Nikolay~Pavlovich Laptev, Ning Dong, Norman Cheng, Oleg Chernoguz, Olivia
  Hart, Omkar Salpekar, Ozlem Kalinli, Parkin Kent, Parth Parekh, Paul Saab,
  Pavan Balaji, Pedro Rittner, Philip Bontrager, Pierre Roux, Piotr Dollar,
  Polina Zvyagina, Prashant Ratanchandani, Pritish Yuvraj, Qian Liang, Rachad
  Alao, Rachel Rodriguez, Rafi Ayub, Raghotham Murthy, Raghu Nayani, Rahul
  Mitra, Rangaprabhu Parthasarathy, Raymond Li, Rebekkah Hogan, Robin Battey,
  Rocky Wang, Russ Howes, Ruty Rinott, Sachin Mehta, Sachin Siby, Sai~Jayesh
  Bondu, Samyak Datta, Sara Chugh, Sara Hunt, Sargun Dhillon, Sasha Sidorov,
  Satadru Pan, Saurabh Mahajan, Saurabh Verma, Seiji Yamamoto, Sharadh
  Ramaswamy, Shaun Lindsay, Shaun Lindsay, Sheng Feng, Shenghao Lin,
  Shengxin~Cindy Zha, Shishir Patil, Shiva Shankar, Shuqiang Zhang, Shuqiang
  Zhang, Sinong Wang, Sneha Agarwal, Soji Sajuyigbe, Soumith Chintala,
  Stephanie Max, Stephen Chen, Steve Kehoe, Steve Satterfield, Sudarshan
  Govindaprasad, Sumit Gupta, Summer Deng, Sungmin Cho, Sunny Virk, Suraj
  Subramanian, Sy~Choudhury, Sydney Goldman, Tal Remez, Tamar Glaser, Tamara
  Best, Thilo Koehler, Thomas Robinson, Tianhe Li, Tianjun Zhang, Tim Matthews,
  Timothy Chou, Tzook Shaked, Varun Vontimitta, Victoria Ajayi, Victoria
  Montanez, Vijai Mohan, Vinay~Satish Kumar, Vishal Mangla, Vlad Ionescu, Vlad
  Poenaru, Vlad~Tiberiu Mihailescu, Vladimir Ivanov, Wei Li, Wenchen Wang,
  Wenwen Jiang, Wes Bouaziz, Will Constable, Xiaocheng Tang, Xiaojian Wu,
  Xiaolan Wang, Xilun Wu, Xinbo Gao, Yaniv Kleinman, Yanjun Chen, Ye~Hu,
  Ye~Jia, Ye~Qi, Yenda Li, Yilin Zhang, Ying Zhang, Yossi Adi, Youngjin Nam,
  Yu~Wang, Yu~Zhao, Yuchen Hao, Yundi Qian, Yunlu Li, Yuzi He, Zach Rait,
  Zachary DeVito, Zef Rosnbrick, Zhaoduo Wen, Zhenyu Yang, Zhiwei Zhao, and
  Zhiyu Ma. 2024.
\newblock \href {http://arxiv.org/abs/2407.21783} {The {Llama} 3 herd of
  models}.
\newblock arXiv preprint arXiv:2407.21783v3.

\bibitem[{Gupta and Belnap(1993)}]{gupta1993revision}
Anil Gupta and Nuel Belnap. 1993.
\newblock \emph{The Revision Theory of Truth}.
\newblock MIT Press, Cambridge, MA.

\bibitem[{Han et~al.(2024)Han, Schoelkopf, Zhao, Qi, Riddell, Zhou, Coady,
  Peng, Qiao, Benson, Sun, Wardle-Solano, Szab{\'o}, Zubova, Burtell, Fan, Liu,
  Wong, Sailor, Ni, Nan, Kasai, Yu, Zhang, Fabbri, Kryscinski, Yavuz, Liu, Lin,
  Joty, Zhou, Xiong, Ying, Cohan, and Radev}]{han2022folio}
Simeng Han, Hailey Schoelkopf, Yilun Zhao, Zhenting Qi, Martin Riddell, Wenfei
  Zhou, James Coady, David Peng, Yujie Qiao, Luke Benson, Lucy Sun, Alexander
  Wardle-Solano, Hannah Szab{\'o}, Ekaterina Zubova, Matthew Burtell, Jonathan
  Fan, Yixin Liu, Brian Wong, Malcolm Sailor, Ansong Ni, Linyong Nan, Jungo
  Kasai, Tao Yu, Rui Zhang, Alexander Fabbri, Wojciech~Maciej Kryscinski, Semih
  Yavuz, Ye~Liu, Xi~Victoria Lin, Shafiq Joty, Yingbo Zhou, Caiming Xiong, Rex
  Ying, Arman Cohan, and Dragomir Radev. 2024.
\newblock \href {https://doi.org/10.18653/v1/2024.emnlp-main.1229} {{FOLIO}:
  {Natural} language reasoning with first-order logic}.
\newblock In \emph{Proceedings of the 2024 Conference on Empirical Methods in
  Natural Language Processing}, pages 22017--22031, Miami, Florida, USA.
  Association for Computational Linguistics.

\bibitem[{Hewitt and Liang(2019)}]{hewitt2019control}
John Hewitt and Percy Liang. 2019.
\newblock \href {https://doi.org/10.18653/v1/D19-1275} {Designing and
  interpreting probes with control tasks}.
\newblock In \emph{Proceedings of the 2019 Conference on Empirical Methods in
  Natural Language Processing and the 9th International Joint Conference on
  Natural Language Processing (EMNLP-IJCNLP)}, pages 2733--2743, Hong Kong,
  China. Association for Computational Linguistics.

\bibitem[{Holm(1979)}]{holm1979simple}
Sture Holm. 1979.
\newblock A simple sequentially rejective multiple test procedure.
\newblock \emph{Scandinavian Journal of Statistics}, 6(2):65--70.

\bibitem[{Koehn(2004)}]{koehn2004statistical}
Philipp Koehn. 2004.
\newblock Statistical significance tests for machine translation evaluation.
\newblock In \emph{Proceedings of the 2004 Conference on Empirical Methods in
  Natural Language Processing}, pages 388--395, Barcelona, Spain. Association
  for Computational Linguistics.

\bibitem[{Kripke(1975)}]{kripke1975outline}
Saul Kripke. 1975.
\newblock \href {https://doi.org/10.2307/2024634} {Outline of a theory of
  truth}.
\newblock \emph{The Journal of Philosophy}, 72(19):690--716.

\bibitem[{Leitgeb(2005)}]{leitgeb2005truth}
Hannes Leitgeb. 2005.
\newblock \href {https://doi.org/10.1007/s10992-004-3758-3} {What truth depends
  on}.
\newblock \emph{Journal of Philosophical Logic}, 34(2):155--192.

\bibitem[{Li et~al.(2023)Li, Patel, Vi{\'e}gas, Pfister, and
  Wattenberg}]{li2023inference}
Kenneth Li, Oam Patel, Fernanda Vi{\'e}gas, Hanspeter Pfister, and Martin
  Wattenberg. 2023.
\newblock \href {https://doi.org/10.52202/075280-1797} {Inference-time
  intervention: {Eliciting} truthful answers from a language model}.
\newblock In \emph{Advances in Neural Information Processing Systems},
  volume~36, pages 41451--41530. Curran Associates, Inc.

\bibitem[{Lin et~al.(2025)Lin, Le~Bras, Richardson, Sabharwal, Poovendran,
  Clark, and Choi}]{lin2025zebralogic}
Bill~Yuchen Lin, Ronan Le~Bras, Kyle Richardson, Ashish Sabharwal, Radha
  Poovendran, Peter Clark, and Yejin Choi. 2025.
\newblock {ZebraLogic}: {On} the scaling limits of {LLMs} for logical
  reasoning.
\newblock In \emph{Proceedings of the 42nd International Conference on Machine
  Learning}, volume 267 of \emph{Proceedings of Machine Learning Research},
  pages 37889--37905. PMLR.

\bibitem[{Marks and Tegmark(2024)}]{marks2024geometry}
Samuel Marks and Max Tegmark. 2024.
\newblock The geometry of truth: {Emergent} linear structure in large language
  model representations of true/false datasets.
\newblock In \emph{First Conference on Language Modeling}.

\bibitem[{McCoy et~al.(2024)McCoy, Yao, Friedman, Hardy, and
  Griffiths}]{mccoy2024embers}
R.~Thomas McCoy, Shunyu Yao, Dan Friedman, Mathew~D. Hardy, and Thomas~L.
  Griffiths. 2024.
\newblock \href {https://doi.org/10.1073/pnas.2322420121} {Embers of
  autoregression show how large language models are shaped by the problem they
  are trained to solve}.
\newblock \emph{Proceedings of the National Academy of Sciences},
  121(41):e2322420121.

\bibitem[{Mizrahi et~al.(2024)Mizrahi, Kaplan, Malkin, Dror, Shahaf, and
  Stanovsky}]{mizrahi2024state}
Moran Mizrahi, Guy Kaplan, Dan Malkin, Rotem Dror, Dafna Shahaf, and Gabriel
  Stanovsky. 2024.
\newblock \href {https://doi.org/10.1162/tacl_a_00681} {State of what art? {A}
  call for multi-prompt {LLM} evaluation}.
\newblock \emph{Transactions of the Association for Computational Linguistics},
  12:933--949.

\bibitem[{Mondorf and Plank(2024)}]{mondorf2024truthquest}
Philipp Mondorf and Barbara Plank. 2024.
\newblock \href {https://doi.org/10.18653/v1/2024.emnlp-main.404} {Liar, liar,
  logical mire: {A} benchmark for suppositional reasoning in large language
  models}.
\newblock In \emph{Proceedings of the 2024 Conference on Empirical Methods in
  Natural Language Processing}, pages 7114--7137, Miami, Florida, USA.
  Association for Computational Linguistics.

\bibitem[{de~Moura and Bj{\o}rner(2008)}]{demoura2008z3}
Leonardo de~Moura and Nikolaj Bj{\o}rner. 2008.
\newblock \href {https://doi.org/10.1007/978-3-540-78800-3_24} {{Z3}: {An}
  efficient {SMT} solver}.
\newblock In \emph{Tools and Algorithms for the Construction and Analysis of
  Systems (TACAS 2008)}, volume 4963 of \emph{Lecture Notes in Computer
  Science}, pages 337--340, Berlin, Heidelberg. Springer.

\bibitem[{Olausson et~al.(2023)Olausson, Gu, Lipkin, Zhang, Solar-Lezama,
  Tenenbaum, and Levy}]{olausson2023linc}
Theo~X. Olausson, Alex Gu, Benjamin Lipkin, Cedegao~E. Zhang, Armando
  Solar-Lezama, Joshua~B. Tenenbaum, and Roger Levy. 2023.
\newblock \href {https://doi.org/10.18653/v1/2023.emnlp-main.313} {{LINC}: {A}
  neurosymbolic approach for logical reasoning by combining language models
  with first-order logic provers}.
\newblock In \emph{Proceedings of the 2023 Conference on Empirical Methods in
  Natural Language Processing}, pages 5153--5176, Singapore. Association for
  Computational Linguistics.

\bibitem[{Pan et~al.(2023)Pan, Albalak, Wang, and Wang}]{pan2023logiclm}
Liangming Pan, Alon Albalak, Xinyi Wang, and William~Yang Wang. 2023.
\newblock \href {https://doi.org/10.18653/v1/2023.findings-emnlp.248}
  {{Logic-LM}: {Empowering} large language models with symbolic solvers for
  faithful logical reasoning}.
\newblock In \emph{Findings of the Association for Computational Linguistics:
  EMNLP 2023}, pages 3806--3824, Singapore. Association for Computational
  Linguistics.

\bibitem[{Poulis et~al.(2026)Poulis, Crovella, and Terzi}]{poulis2026limits}
Angelos Poulis, Mark Crovella, and Evimaria Terzi. 2026.
\newblock \href {https://aclanthology.org/2026.blackboxnlp-1.26/} {Testing the
  limits of truth directions in {LLM}s}.
\newblock In \emph{Proceedings of the 9th {B}lackbox{NLP} Workshop: Analyzing
  and Interpreting Neural Networks for {NLP}}, pages 419--447, Budapest,
  Hungary. Association for Computational Linguistics.

\bibitem[{Rabern et~al.(2013)Rabern, Rabern, and
  Macauley}]{rabern2013dangerous}
Landon Rabern, Brian Rabern, and Matthew Macauley. 2013.
\newblock \href {https://doi.org/10.1007/s10992-012-9246-2} {Dangerous
  reference graphs and semantic paradoxes}.
\newblock \emph{Journal of Philosophical Logic}, 42(5):727--765.

\bibitem[{Razeghi et~al.(2022)Razeghi, Logan~IV, Gardner, and
  Singh}]{razeghi2022impact}
Yasaman Razeghi, Robert~L. Logan~IV, Matt Gardner, and Sameer Singh. 2022.
\newblock \href {https://doi.org/10.18653/v1/2022.findings-emnlp.59} {Impact of
  pretraining term frequencies on few-shot numerical reasoning}.
\newblock In \emph{Findings of the Association for Computational Linguistics:
  EMNLP 2022}, pages 840--854, Abu Dhabi, United Arab Emirates. Association for
  Computational Linguistics.

\bibitem[{Roy and Vetterli(2007)}]{roy2007effective}
Olivier Roy and Martin Vetterli. 2007.
\newblock The effective rank: {A} measure of effective dimensionality.
\newblock In \emph{Proceedings of the 15th European Signal Processing
  Conference}, pages 606--610, Poznan, Poland. EURASIP.

\bibitem[{Saparov and He(2023)}]{saparov2023greedy}
Abulhair Saparov and He~He. 2023.
\newblock Language models are greedy reasoners: {A} systematic formal analysis
  of chain-of-thought.
\newblock In \emph{The Eleventh International Conference on Learning
  Representations}.

\bibitem[{Sclar et~al.(2024)Sclar, Choi, Tsvetkov, and
  Suhr}]{sclar2024quantifying}
Melanie Sclar, Yejin Choi, Yulia Tsvetkov, and Alane Suhr. 2024.
\newblock Quantifying language models' sensitivity to spurious features in
  prompt design or: {How} {I} learned to start worrying about prompt
  formatting.
\newblock In \emph{The Twelfth International Conference on Learning
  Representations}.

\bibitem[{Tafjord et~al.(2021)Tafjord, Dalvi, and
  Clark}]{tafjord2021proofwriter}
Oyvind Tafjord, Bhavana Dalvi, and Peter Clark. 2021.
\newblock \href {https://doi.org/10.18653/v1/2021.findings-acl.317}
  {{ProofWriter}: {Generating} implications, proofs, and abductive statements
  over natural language}.
\newblock In \emph{Findings of the Association for Computational Linguistics:
  ACL-IJCNLP 2021}, pages 3621--3634, Online. Association for Computational
  Linguistics.

\bibitem[{Thrush et~al.(2024)Thrush, Moore, Monares, Potts, and
  Kiela}]{thrush2024strange}
Tristan Thrush, Jared Moore, Miguel Monares, Christopher Potts, and Douwe
  Kiela. 2024.
\newblock \href {https://doi.org/10.18653/v1/2024.acl-long.482} {{I} am a
  {Strange Dataset}: {Metalinguistic} tests for language models}.
\newblock In \emph{Proceedings of the 62nd Annual Meeting of the Association
  for Computational Linguistics (Volume 1: Long Papers)}, pages 8888--8907,
  Bangkok, Thailand. Association for Computational Linguistics.

\bibitem[{Tian et~al.(2021)Tian, Li, Chen, Xiao, He, and
  Jin}]{tian2021logicnli}
Jidong Tian, Yitian Li, Wenqing Chen, Liqiang Xiao, Hao He, and Yaohui Jin.
  2021.
\newblock \href {https://doi.org/10.18653/v1/2021.emnlp-main.303} {Diagnosing
  the first-order logical reasoning ability through {LogicNLI}}.
\newblock In \emph{Proceedings of the 2021 Conference on Empirical Methods in
  Natural Language Processing}, pages 3738--3747, Online and Punta Cana,
  Dominican Republic. Association for Computational Linguistics.

\bibitem[{Tishby and Zaslavsky(2015)}]{tishby2015deep}
Naftali Tishby and Noga Zaslavsky. 2015.
\newblock \href {https://doi.org/10.1109/ITW.2015.7133169} {Deep learning and
  the information bottleneck principle}.
\newblock In \emph{2015 IEEE Information Theory Workshop (ITW)}, pages 1--5,
  Jerusalem, Israel. IEEE.

\bibitem[{Walicki(2009)}]{walicki2009reference}
Micha{\l} Walicki. 2009.
\newblock \href {https://doi.org/10.1007/s11229-008-9392-9} {Reference,
  paradoxes and truth}.
\newblock \emph{Synthese}, 171(1):195--226.

\bibitem[{Wei et~al.(2025)Wei, Wu, Wan, Suresh, Tan, Zhou, Koyejo, Wang, and
  Aiken}]{wei2025satbench}
Anjiang Wei, Yuheng Wu, Yingjia Wan, Tarun Suresh, Huanmi Tan, Zhanke Zhou,
  Sanmi Koyejo, Ke~Wang, and Alex Aiken. 2025.
\newblock \href {https://doi.org/10.18653/v1/2025.emnlp-main.1716} {{SATBench}:
  {Benchmarking} {LLMs}' logical reasoning via automated puzzle generation from
  {SAT} formulas}.
\newblock In \emph{Proceedings of the 2025 Conference on Empirical Methods in
  Natural Language Processing}, pages 33832--33849, Suzhou, China. Association
  for Computational Linguistics.

\bibitem[{Xie et~al.(2025{\natexlab{a}})Xie, Huang, Zhang, Yu, Chen, Lin, Li,
  Ghazi, and Kumar}]{xie2024memorization}
Chulin Xie, Yangsibo Huang, Chiyuan Zhang, Da~Yu, Xinyun Chen, Bill~Yuchen Lin,
  Bo~Li, Badih Ghazi, and Ravi Kumar. 2025{\natexlab{a}}.
\newblock \href {https://doi.org/10.18653/v1/2025.ijcnlp-long.148} {On
  memorization of large language models in logical reasoning}.
\newblock In \emph{Proceedings of the 14th International Joint Conference on
  Natural Language Processing and the 4th Conference of the Asia-Pacific
  Chapter of the Association for Computational Linguistics}, pages 2742--2785,
  Mumbai, India. The Asian Federation of Natural Language Processing and The
  Association for Computational Linguistics.

\bibitem[{Xie et~al.(2025{\natexlab{b}})Xie, Gao, Ren, Luo, Hong, Dai, Zhou,
  Qiu, Wu, and Luo}]{xie2025logicrl}
Tian Xie, Zitian Gao, Qingnan Ren, Haoming Luo, Yuqian Hong, Bryan Dai, Joey
  Zhou, Kai Qiu, Zhirong Wu, and Chong Luo. 2025{\natexlab{b}}.
\newblock \href {http://arxiv.org/abs/2502.14768} {{Logic-RL}: {Unleashing}
  {LLM} reasoning with rule-based reinforcement learning}.
\newblock arXiv preprint arXiv:2502.14768v1.

\bibitem[{Yablo(1993)}]{yablo1993paradox}
Stephen Yablo. 1993.
\newblock \href {https://doi.org/10.1093/analys/53.4.251} {Paradox without
  self-reference}.
\newblock \emph{Analysis}, 53(4):251--252.

\bibitem[{Yang et~al.(2025)Yang, Li, Yang, Zhang, Hui, Zheng, Yu, Gao, Huang,
  Lv, Zheng, Liu, Zhou, Huang, Hu, Ge, Wei, Lin, Tang, Yang, Tu, Zhang, Yang,
  Yang, Zhou, Zhou, Lin, Dang, Bao, Yang, Yu, Deng, Li, Xue, Li, Zhang, Wang,
  Zhu, Men, Gao, Liu, Luo, Li, Tang, Yin, Ren, Wang, Zhang, Ren, Fan, Su,
  Zhang, Zhang, Wan, Liu, Wang, Cui, Zhang, Zhou, and Qiu}]{qwen3}
An~Yang, Anfeng Li, Baosong Yang, Beichen Zhang, Binyuan Hui, Bo~Zheng, Bowen
  Yu, Chang Gao, Chengen Huang, Chenxu Lv, Chujie Zheng, Dayiheng Liu, Fan
  Zhou, Fei Huang, Feng Hu, Hao Ge, Haoran Wei, Huan Lin, Jialong Tang, Jian
  Yang, Jianhong Tu, Jianwei Zhang, Jianxin Yang, Jiaxi Yang, Jing Zhou,
  Jingren Zhou, Junyang Lin, Kai Dang, Keqin Bao, Kexin Yang, Le~Yu, Lianghao
  Deng, Mei Li, Mingfeng Xue, Mingze Li, Pei Zhang, Peng Wang, Qin Zhu, Rui
  Men, Ruize Gao, Shixuan Liu, Shuang Luo, Tianhao Li, Tianyi Tang, Wenbiao
  Yin, Xingzhang Ren, Xinyu Wang, Xinyu Zhang, Xuancheng Ren, Yang Fan, Yang
  Su, Yichang Zhang, Yinger Zhang, Yu~Wan, Yuqiong Liu, Zekun Wang, Zeyu Cui,
  Zhenru Zhang, Zhipeng Zhou, and Zihan Qiu. 2025.
\newblock \href {http://arxiv.org/abs/2505.09388} {{Qwen3} technical report}.
\newblock arXiv preprint arXiv:2505.09388v1.

\bibitem[{Yang et~al.(2024)Yang, Gribovskaya, Kassner, Geva, and
  Riedel}]{yang2024latent}
Sohee Yang, Elena Gribovskaya, Nora Kassner, Mor Geva, and Sebastian Riedel.
  2024.
\newblock \href {https://doi.org/10.18653/v1/2024.acl-long.550} {Do large
  language models latently perform multi-hop reasoning?}
\newblock In \emph{Proceedings of the 62nd Annual Meeting of the Association
  for Computational Linguistics (Volume 1: Long Papers)}, pages 10210--10229,
  Bangkok, Thailand. Association for Computational Linguistics.

\bibitem[{Ye et~al.(2023)Ye, Chen, Dillig, and Durrett}]{ye2023satlm}
Xi~Ye, Qiaochu Chen, Isil Dillig, and Greg Durrett. 2023.
\newblock \href {https://doi.org/10.52202/075280-1974} {{SatLM}:
  {Satisfiability}-aided language models using declarative prompting}.
\newblock In \emph{Advances in Neural Information Processing Systems},
  volume~36, pages 45548--45580. Curran Associates, Inc.

\bibitem[{Yeh(2000)}]{yeh2000accurate}
Alexander Yeh. 2000.
\newblock \href {https://doi.org/10.3115/992730.992783} {More accurate tests
  for the statistical significance of result differences}.
\newblock In \emph{COLING 2000 Volume 2: The 18th International Conference on
  Computational Linguistics}, pages 947--953, Saarbr{\"u}cken, Germany.
  Association for Computational Linguistics.

\bibitem[{Zhang et~al.(2026)Zhang, Chen, and Chen}]{zhang2026matchedpair}
Leizhen Zhang, Shuhan Chen, and Sheng Chen. 2026.
\newblock \href {https://doi.org/10.1145/3808209} {Satisfiability solving with
  {LLMs}: {A} matched-pair evaluation of reasoning capability}.
\newblock \emph{Proceedings of the ACM on Software Engineering},
  3(FSE):4599--4621.

\end{thebibliography}
\bibliographystyle{acl_natbib}

\clearpage
\appendix
\renewcommand{\dbltopfraction}{1}\setcounter{dbltopnumber}{4}\setlength{\dblfloatsep}{8pt plus 2pt minus 2pt}\setlength{\abovecaptionskip}{6pt}\renewcommand{\dblfloatpagefraction}{0.85}
\uchyph=0

\section{Experimental Details}

\paragraph{Models.}
We used four models: Gemma-2-9B-it, Llama-3.2-11B-Vision-Instruct, Llama-3.3-70B-Instruct, and Qwen3-VL-8B-Instruct. For all four models, we loaded the public weights from Hugging Face at fixed repository revisions in BF16 without quantization. Because we provided text-only input to Llama-11B, its eight cross-attention layers were not executed, and we numbered the layers and computed relative depths using only the 32 executed self-attention layers. For Llama-70B, we used a third-party re-upload (by unsloth) whose weights, tokenizer, and chat template are the same as those of the official repository.

\paragraph{Software and hardware.}
We used PyTorch and the Transformers library with eager attention. The repository revisions and software versions will be listed with the released code. The three smaller models each ran on a single NVIDIA RTX PRO 6000 GPU (96~GB), and Llama-70B ran on two such GPUs.

\paragraph{Decoding and parsing.}
We generated all responses with greedy decoding (up to 4,096 new tokens), using each model's own chat template. We formed left-padded batches of 16 rows (Gemma and Llama-70B) or 32 rows (Llama-11B and Qwen) by input length, regardless of condition. We treated a response as parsed only if it contained exactly one ``FINAL: LABEL'' line and ended with that line. The parser ignored letter case and allowed Markdown emphasis and a final period or exclamation mark. We also rescored all responses with a stricter parser that allows neither emphasis nor trailing punctuation (Table~4). The analysis in §5 that adjusts for likelihood and length uses a linear probability model fitted separately for each language model. It regresses an indicator of whether the answer is \inc{} on condition, mean body NLL, the log of the number of body tokens, and wording. The estimate reported in §5 averages the four models' condition coefficients with equal weights, and its confidence interval is obtained by a family-level bootstrap (2,000 draws).

\paragraph{Order of operations and control selection.}
For each model, we proceeded in five steps. (1) We computed input likelihoods by teacher forcing, in left-padded batches of 8 rows (Gemma and \mbox{Llama-70B}) or 16 rows (Llama-11B and Qwen), for all rows of the main dataset and of the ordinary constraint problems, the grounded variants (§3), and the natural-language items. The likelihoods are mean negative log-likelihoods per token over two spans: the item body up to and including the question sentence (body) and the entire user prompt (content). The difference $\Delta \mathrm{NLL}$ used when selecting a control is the value for the circular-system row minus the value for the candidate variant. (2) Among the variants within the tolerances stated in §3, we selected as the control the one with the smallest $|\Delta \mathrm{NLL}_{\mathrm{body}}|+0.2\,|\log(T_{\mathrm{src}}/T_{\mathrm{var}})|$. (3) We generated all responses. (4) We reran 16 rows, chosen in advance by a hash of their identifiers, at batch size one. (5) We ran one teacher-forcing pass per row to record all hidden states and one input-only pass per row to record the attention views over the prompt. After the main analysis, we reran 64 rows per model at batch size one (§5).

\paragraph{Teacher-forcing recomputation.}
Rather than recording hidden states during batched generation, we recomputed them by running one teacher-forcing pass per row at batch size one. In an FP32 check on randomly initialized small models of all four architectures, this procedure reproduced the states from batch-size-one generation to within $9.6\times 10^{-7}$ (absolute error on prompt tokens) and $5.7\times 10^{-7}$ (relative error on generated tokens).

\section{Prompts}

\paragraph{Full prompt text.}
Each prompt is passed as a single user message with no system message; the chat templates of both Llama models add their default system header, in which we set the date to 09 Oct 2026. The user turn begins with the classification rule ``Use classical Boolean values, with no third truth value. Consider ALL simultaneous assignments that satisfy EVERY condition below. Classify the queried item as TRUE if at least one assignment exists and it is true in every satisfying assignment; FALSE if at least one exists and it is false in every satisfying assignment; UNDERDETERMINED if satisfying assignments exist with both query values; INCONSISTENT if no simultaneous assignment satisfies all conditions. Multiple assignments do not by themselves make the query UNDERDETERMINED. All item names are arbitrary labels without extra meaning.'' A blank line separates the classification rule from the item body, which ends with the question sentence. A second blank line separates the item body from the final request ``Briefly explain your reasoning. End with exactly one line in the form FINAL: LABEL, where LABEL is TRUE, FALSE, UNDERDETERMINED, or INCONSISTENT.'' Both quoted passages are verbatim.

\section{Item Construction and Internal Analysis Details}

\paragraph{Cycle removal.}
Replacing cycle-closing references with constants in a grounded twin (§2) breaks every cycle. The procedure replaces every reference to the utterance containing it or to a later utterance in the random order, so it may replace more references than necessary to remove all cycles. In Figure~1 (order C, D, A, B), D's reference to A is also replaced with a constant, but after C's references are replaced with constants, D's reference to A is no longer part of a cycle. All 144 circular systems are strongly connected: every utterance depends on every other utterance, either directly or through other utterances. The systems are pairwise distinct as rooted Boolean systems when operators and literal signs are included. Retaining only the edges and the root gives 136 distinct rooted dependency graphs, 55 of which appear in the 56 core families. The number of references replaced per family ranges from 3 to 9 (median 5, 697 in total). Within each family, the constants come from a single assignment. In the 56 core families, this is the unique consistent assignment; in the 52 families with multiple consistent assignments, it is one of those assignments (so the twin preserves one solution); and in the 36 families with no consistent assignment, it is an assignment that does not satisfy the circular system. Because every grounded twin's dependency graph is acyclic, its consistent assignment is unique. Thus, outside the core families, the twin's correct answer is \tru{} or \fal{} (18 \tru{} and 18 \fal{} among the 36 \und{} families, and likewise among the 36 \inc{} families). The grounded twin's queried utterance is settled in the first round in 27 of the 56 core families (in 21, the queried utterance mentions only constants, and in 6, a single constant settles the value of the connective), and in 2--4 rounds in the remaining 29 families. These two groups are the one-step families and propagation families of §6 (post hoc split). The 95\% confidence interval for the difference in the behavioral contrast between the two groups was obtained using a family-level bootstrap (10,000 draws), with families resampled independently within each group. We obtained the $p$-value by permuting the group labels (20,000 draws).

\paragraph{Truth and resolution scores.}
We extract residual states at two positions, the last token of the item body (the end of the question sentence) and the last token of the prompt; all reported results use the former. The truth direction is the difference $w$ between the class means computed on the 128 arithmetic calibration rows. We validate it using eightfold stratified cross-validation, with a validation threshold of 0.90 and 1,000 label permutations ($p\leq 0.001$ at every validated layer). Cross-validated AUCs, here and for the classifier, are computed once on the pooled out-of-fold scores. Because folds split true--false minimal pairs, the AUC on propositions not used to fit the direction can be below 0.5 at layers that lack a truth signal. The truth score of a state $h$ is $(h^Tw-b)/s$, where $b$ is the midpoint between the two projected class means and $s$ is the pooled within-class standard deviation of the projections. We estimate both using all 128 rows; cross-validation is used only to validate layers. The resolution score is the truth score multiplied by $+1$ when the correct answer is \tru{} and by $-1$ when it is \fal{}.

\paragraph{Late-layer update and attention rank.}
With $h_l$ the state after layer $l$ at the question's last token ($1\leq l\leq L$; $h_L$ is the state after final normalization), the late-layer update (IH-U) is the mean of $\lVert h_{l+1}-h_l\rVert/\lVert h_l\rVert$ over the last third (rounded up) of the $L-1$ transitions. This calculation includes the final step into $h_L$ (the last layer and final normalization). An unplanned sensitivity analysis omits this one step from the same transitions. Attention rank (IH-A) is computed in an input-only forward pass as the effective rank (\citealp{roy2007effective}) of the head-averaged attention from the last 32 tokens of the prompt (the end of the response instruction and the chat template, all following the question) to all tokens of the prompt, divided by 32, at relative depths 0.1, 0.3, 0.5, 0.7, and 0.9, namely (counting from layer 1) layers 5, 13, 21, 30, and 38 of 42 in Gemma; layers 4, 10, 17, 23, and 29 of 32 in Llama-11B; layers 9, 25, 41, 56, and 72 of 80 in Llama-70B; and layers 5, 11, 19, 25, and 33 of 36 in Qwen. IH-A uses the value at depth 0.5, whereas the classifier uses the values at all five depths.

\paragraph{Tests and length adjustment.}
The primary internal tests use a family-level sign-flip randomization test (100,000 draws) with Holm correction across IH-R, IH-U, and IH-A. For length adjustment, we fit separate least-squares regressions for circular systems and grounded twins within each model and wording. In each regression, the measure is regressed across families on the log of the number of input tokens. We average the two slopes, multiply this average by each pair's difference in log length, and subtract the product from that pair's difference in the measure. In the randomization and permutation tests, the $p$-value is $(k+1)/(B+1)$, where $B$ is the number of draws and $k$ is the number of draws in which the statistic is at least as extreme as the observed value. Sign-flip and group-label tests are two-sided; the AUC permutation tests count only draws at least as high as the observed value. Bootstrap intervals are percentile intervals. We used fixed seeds for all random draws.

\paragraph{Classifier.}
The 15 classification features are the resolution scores averaged over the validated layers in each of five contiguous bins of roughly equal size into which the $L$ layers are divided (bins without validated layers are set to 0: the first bin in all models and the second bin in Gemma and Qwen), the residual updates $\lVert h_{l+1}-h_l\rVert/\lVert h_l\rVert$ averaged over five similarly divided bins of the $L-1$ transitions, and the attention ranks at the five depths. Before splitting into folds, each feature was standardized without labels using the mean and standard deviation across the 224 circular-system and grounded-twin rows of that model (56 core families, two wordings). For pairwise classification, we place each pair's members in random order and use L2-regularized logistic regression without an intercept. We select the regularization strength from \{0.01, 0.1, 1, 10\} using inner-fold accuracy (ties go to the smallest value), with folds grouped by family. We run 1,000 family-level permutations, repeating the entire procedure for each permutation. During training for the leave-one-model-out test (LOMO), folds and permutations are grouped by model--family combination. Each permutation flips only the training labels and reselects the regularization strength. The held-out model's pairs come from the same 56 families as the training pairs. In the prespecified small-subset test, we computed permutation importance (ROC AUC, 50 repeats) on the training models in each split, reselected the regularization strength using only the top three features, and refit the classifier.

\paragraph{Feature scaling.}
The standardization described above does not use labels, but it is transductive because it uses the distribution of the test rows. Specifically, each feature's standard deviation is computed across all 224 rows of that model, including those later used for testing. Taking the within-pair difference cancels the mean. In a post hoc check (Table~8), all scales are refit using only the training rows in every outer and inner fold. In its leave-one-model-out part, the scales pool the training models, the held-out model's rows use the same pooled statistics, and the inner folds are grouped by family across models. The joint holdout test in this check holds out one model together with one of the eight family folds from training, computes the scales across all remaining training-model rows, and groups the inner folds by family across models. Its permutation test (1,000 draws) flips the training labels at the family level, sharing the same flips across the two wordings, models, and outer folds. We reselect the regularization strength on each draw. The resulting $p$-values are Holm-adjusted across the four held-out models. This check is exploratory.

\paragraph{Other analyses.}
We compared the late-layer updates and attention ranks of circular-system rows answered \inc{} with those of the remaining circular-system rows (results in Table~5). Figure~3 shows the resolution score profiles as within-family differences. The constraint-twin curve compares items in different formats, so a shift in truth score common to both answers does not cancel. The curve reflects this shift through the 31:25 split of correct answer labels. The natural-language comparison is presented in Table~6. After inspecting an interim transfer analysis conducted without Llama-70B, we also fit a classifier on the resolution score features alone; it reached AUC 0.70--0.75 on every held-out model, but no transfer was significant after Holm correction ($p\geq 0.16$).

\section{Additional Results}

Tables~3--8 break down the main-text results by model or supplement them.

\setcounter{table}{2}
\begin{table*}[t]
\centering\small
\setlength{\tabcolsep}{3.5pt}
\begin{adjustbox}{max width=\textwidth}
\begin{tabular}{@{}lrrrrrrrrrrrrr@{}}
\toprule
 & \multicolumn{4}{c}{Circular systems} & \multicolumn{4}{c}{Grounded twins} & \multicolumn{4}{c}{Constraint twins} & Matched\\
\cmidrule(lr){2-5}\cmidrule(lr){6-9}\cmidrule(lr){10-13}
Model & $n$ & \textsc{inc} & Correct & Unparsed & $n$ & \textsc{inc} & Correct & Unparsed & $n$ & \textsc{inc} & Correct & Unparsed & controls\\
\midrule
Gemma-2-9B & 112 & 69 & 19 & 1 & 112 & 5 & 79 & 0 & 56 & 2 & 7 & 1 & 131/144\\
Llama-3.2-11B & 112 & 62 & 18 & 8 & 112 & 49 & 42 & 0 & 56 & 13 & 11 & 7 & 131/144\\
Llama-3.3-70B & 112 & 58 & 34 & 0 & 112 & 9 & 93 & 0 & 56 & 2 & 5 & 0 & 141/144\\
Qwen3-VL-8B & 112 & 14 & 72 & 18 & 112 & 1 & 110 & 1 & 56 & 3 & 17 & 15 & 26/144\\
\bottomrule
\end{tabular}
\end{adjustbox}
\caption{Response counts by model for the 56 core families. For circular systems and grounded twins, counts combine both wordings; \textsc{inc}: \inc{} answers. Matched controls: circular truth-wording rows in all 144 families for which the likelihood rule found a perplexity-matched grounded control before generation.}\label{tab:app-counts}
\end{table*}

\setcounter{table}{3}
\begin{table*}[t]
\centering\small
\setlength{\tabcolsep}{3pt}
\begin{adjustbox}{max width=\textwidth}
\begin{tabular}{@{}lrrrrrrrrrrrrr@{}}
\toprule
 & \multicolumn{2}{c}{Specificity} & \multicolumn{2}{c}{Batch size 1} & & \multicolumn{2}{c}{108 satisfiable} & \multicolumn{3}{c}{Among incorrect or unparsed} & \multicolumn{3}{c@{}}{Ordinary constraint problems}\\
\cmidrule(lr){2-3}\cmidrule(lr){4-5}\cmidrule(lr){7-8}\cmidrule(lr){9-11}\cmidrule(l){12-14}
Model & pp & Holm $p$ & 16 rows & 64 rows & Limit & Circular & Grounded & Circular & Grounded & Constraint & Sat. (24) & Unsat. (8) & Correct (32)\\
\midrule
Gemma-2-9B & +58.0 & $4\times10^{-5}$ & 14/16 & 54/64 & 0/818 & 129 & 11 & 69/93 & 5/33 & 2/49 & 0 & 0 & 10\\
Llama-3.2-11B & +32.1 & $6.6\times10^{-4}$ & 7/16 & 38/64 & 42/817 & 113 & 101 & 62/94 & 49/70 & 13/45 & 6 & 0 & 7\\
Llama-3.3-70B & +48.2 & $4\times10^{-5}$ & 10/16 & 37/64 & 0/821 & 111 & 21 & 58/78 & 9/19 & 2/51 & 0 & 0 & 12\\
Qwen3-VL-8B & +7.1 & 0.189 & 10/16 & 46/64 & 90/830 & 20 & 5 & 14/40 & 1/2 & 3/39 & 1 & 3 & 21\\
\bottomrule
\end{tabular}
\end{adjustbox}
\caption{Specificity contrast by model (circular system minus constraint twin, in percentage points; Holm-adjusted across models; per-model tests were not prespecified), counts of final labels reproduced by the batch-size-one reruns of §5.3 (two unparsed responses count as a match), counts of rows that reached the 4,096-token limit, and counts of \inc{} responses for the secondary outcomes. 108 satisfiable: counts for these circular systems and their grounded twins (both wordings; of 216 each). Among incorrect or unparsed: in the 56 core families (a descriptive comparison conditioned on error). Ordinary constraint problems: counts of \inc{} responses on satisfiable and unsatisfiable items, and counts of correct answers. With the stricter parser (Appendix~A), the number of unparsed rows rises from 148 to 229 of 3,286; the primary contrast becomes +29.5 percentage points (95\% CI [24.3, 34.6]) and the specificity contrast becomes +34.8 percentage points ([28.8, 40.8]), both with Holm-adjusted $p=2\times 10^{-5}$.}\label{tab:app-secondary}
\end{table*}

\setcounter{table}{4}
\begin{table*}[t]
\centering\small
\setlength{\tabcolsep}{3.5pt}
\begin{adjustbox}{max width=\textwidth}
\begin{tabular}{@{}lllrrrrrr@{}}
\toprule
 & & & & \multicolumn{5}{c}{Leave-one-model-out test (LOMO)}\\
\cmidrule(lr){5-9}
Model & Validated layer range & Best layer (AUC) & CV AUC & AUC & Accuracy & Permutation $p$ & Holm $p$ & Top-3 AUC\\
\midrule
Gemma-2-9B & 20--42 & 30 (0.990) & 0.98 & 0.05 & 0.13 & 0.992 & 1.000 & 0.03\\
Llama-3.2-11B & 14--32 & 16 (0.977) & 1.00 & 0.32 & 0.38 & 0.684 & 1.000 & 0.31\\
Llama-3.3-70B & 21--80 & 47 (0.983) & 1.00 & 0.98 & 0.96 & 0.009 & 0.036 & 0.98\\
Qwen3-VL-8B & 18--36 & 26 (0.985) & 1.00 & 0.96 & 0.88 & 0.021 & 0.063 & 0.94\\
\bottomrule
\end{tabular}
\end{adjustbox}
\caption{Internal results by model. Validated layer range: layers at which the arithmetic truth direction at the question's last token reaches AUC $\geq$ 0.90 (Appendix~C); best layer: the layer with the highest AUC on unseen arithmetic propositions (Figure~3a), the deepest such layer when several tie. CV AUC: within-model nested cross-validation; LOMO: a classifier trained on the other three models and tested on this one (Holm-adjusted across the four held-out models); Top-3 AUC: AUC after refitting the classifier using only the three most important features. For circular systems, differences in late-layer updates and attention ranks between rows answered \inc{} and the remaining rows were at most 0.003 in every model (resolution scores: §7).}\label{tab:app-internal}
\end{table*}

\setcounter{table}{5}
\begin{table*}[t]
\centering\footnotesize
\setlength{\tabcolsep}{4pt}\renewcommand{\arraystretch}{0.88}
\begin{adjustbox}{max width=\textwidth}
\begin{tabular}{@{}lrrrrrr@{}}
\toprule
 & \multicolumn{2}{c}{IH-R} & \multicolumn{2}{c}{IH-U ($\times10^{3}$)} & \multicolumn{2}{c@{}}{IH-A ($\times10^{3}$)}\\
\cmidrule(lr){2-3}\cmidrule(lr){4-5}\cmidrule(l){6-7}
Comparison & Estimate [CI] & $p$ & Estimate [CI] & $p$ & Estimate [CI] & $p$\\
\midrule
Grounded twins & +0.31 [0.16, 0.47] & $2.6\times10^{-4}$ & +0.11 [$-$0.50, 0.73] & 0.733 & $-$0.94 [$-$1.35, $-$0.56] & $6\times10^{-5}$\\
Grounded twins, length-adjusted & +0.27 [0.12, 0.43] & $8.5\times10^{-4}$ & +0.02 [$-$0.59, 0.64] & 0.940 & $-$0.88 [$-$1.29, $-$0.50] & $3\times10^{-5}$\\
Perplexity-matched controls & +0.29 [0.14, 0.45] & $6.1\times10^{-4}$ & +1.77 [0.95, 2.60] & $9\times10^{-5}$ & $-$0.79 [$-$1.35, $-$0.28] & 0.0045\\
Constraint twins & +0.09 [$-$0.04, 0.23] & 0.209 & $-$4.95 [$-$5.94, $-$3.92] & $<10^{-5}$ & +2.52 [2.03, 2.98] & $<10^{-5}$\\
\addlinespace
\multicolumn{7}{@{}l}{\emph{Grounded twins: one-step and propagation families (post hoc analysis, unadjusted)}}\\
\quad One-step families (27) & +0.61 [0.38, 0.86] & $3\times10^{-5}$ & +0.33 [$-$0.72, 1.38] & 0.552 & $-$1.27 [$-$1.89, $-$0.67] & $2.6\times10^{-4}$\\
\quad Propagation families (29) & +0.03 [$-$0.11, 0.18] & 0.697 & $-$0.09 [$-$0.78, 0.58] & 0.791 & $-$0.64 [$-$1.15, $-$0.17] & 0.015\\
\addlinespace
\multicolumn{7}{@{}l}{\emph{Grounded twins by model (Holm-adjusted across models)}}\\
\quad Gemma-2-9B & +0.67 [0.30, 1.07] & 0.0015 & +0.25 [$-$0.90, 1.44] & 0.682 & $-$2.02 [$-$3.11, $-$0.98] & 0.0015\\
\quad Llama-3.2-11B & +0.17 [0.08, 0.27] & 0.0015 & $-$1.76 [$-$3.23, $-$0.36] & 0.085 & +1.40 [0.87, 1.90] & $6\times10^{-5}$\\
\quad Llama-3.3-70B & +0.21 [0.11, 0.32] & $6.3\times10^{-4}$ & +1.35 [0.22, 2.46] & 0.085 & $-$0.07 [$-$0.11, $-$0.02] & 0.0082\\
\quad Qwen3-VL-8B & +0.18 [0.09, 0.28] & $3.2\times10^{-4}$ & +0.59 [$-$0.15, 1.32] & 0.238 & $-$3.08 [$-$4.22, $-$2.01] & $4\times10^{-5}$\\
\addlinespace
\multicolumn{7}{@{}l}{\emph{Natural-language dataset, exploratory analysis (not tested)}}\\
\quad Model average & $-$0.38 & -- & +2.78 & -- & $-$1.12 & --\\
\quad Gemma-2-9B & $-$1.41 & -- & +14.08 & -- & $-$0.63 & --\\
\quad Llama-3.2-11B & +0.18 & -- & +1.23 & -- & $-$0.10 & --\\
\quad Llama-3.3-70B & $-$0.29 & -- & $-$6.06 & -- & $-$0.34 & --\\
\quad Qwen3-VL-8B & $-$0.01 & -- & +1.85 & -- & $-$3.42 & --\\
\bottomrule
\end{tabular}
\end{adjustbox}
\caption{All planned estimates for the three internal hypothesis (IH) tests on the 56 core families, the post hoc split of the grounded-twin comparison (§6.4), and the exploratory natural-language comparison. IH-R (P2, resolution score): comparison item minus circular system (calibration standard deviations); IH-U (P3, late-layer update) and IH-A (P4, attention rank): circular system minus comparison item. Positive values indicate the direction predicted by P2--P4 (constraint twins: P5). Brackets: 95\% CI (family bootstrap); $p$: $p$-value from the family-level sign-flip test, Holm-adjusted across the three tests in the first row and across models in the per-model rows, and unadjusted elsewhere. Perplexity-matched controls: truth wording and models with a matched control; constraint twins: truth wording. In the natural-language dataset (16 wording families), ungrounded self-reference replaces the circular system. In this comparison, the correct answer is confounded with the connective (\tru{} items open ``At least one is the case'', \fal{} items ``Both are the case''), and the unsigned truth score was higher for the grounded item in all 16 pairs in all models. The sign-aligned difference therefore cannot isolate resolution.}\label{tab:app-repeats}
\end{table*}

\setcounter{table}{6}
\begin{table*}[t]
\centering\footnotesize
\setlength{\tabcolsep}{3pt}
\begin{adjustbox}{max width=\textwidth}
\begin{tabular}{@{}lrrr@{\hspace{14pt}}lrrr@{\hspace{14pt}}lrrr@{}}
\toprule
\multicolumn{4}{@{}c}{Attention rank} & \multicolumn{4}{c}{Resolution score} & \multicolumn{4}{c@{}}{Residual update}\\
\cmidrule(r){1-4}\cmidrule(lr){5-8}\cmidrule(l){9-12}
Depth & $d_z$ & $p$ & Holm $p$ & Bin & $d_z$ & $p$ & Holm $p$ & Bin & $d_z$ & $p$ & Holm $p$\\
\midrule
0.1 & +0.48 & $7.5\times10^{-4}$ & 0.0045 & 1 & -- & 1.000 & 1.000 & 1 & +0.90 & $<10^{-5}$ & $1.5\times10^{-4}$\\
0.3 & $-$2.52 & $<10^{-5}$ & $1.5\times10^{-4}$ & 2 & $-$0.40 & 0.0029 & 0.012 & 2 & +0.62 & $4\times10^{-5}$ & $4\times10^{-4}$\\
0.5 & $-$0.62 & $<10^{-5}$ & $1.5\times10^{-4}$ & 3 & $-$0.48 & $2.1\times10^{-4}$ & 0.0017 & 3 & $-$0.77 & $<10^{-5}$ & $1.5\times10^{-4}$\\
0.7 & $-$0.87 & $<10^{-5}$ & $1.5\times10^{-4}$ & 4 & $-$0.56 & $5\times10^{-5}$ & $4.5\times10^{-4}$ & 4 & $-$0.47 & $9.8\times10^{-4}$ & 0.0049\\
0.9 & $-$0.25 & 0.072 & 0.144 & 5 & $-$0.49 & $5.1\times10^{-4}$ & 0.0036 & 5 & +0.38 & 0.0069 & 0.021\\
\bottomrule
\end{tabular}
\end{adjustbox}
\caption{The 15 classifier features tested individually on the 56 core families. For each family, the feature difference (circular system minus grounded twin) is computed before within-model standardization and averaged over the two wordings and four models. $d_z$: mean family-level difference divided by its standard deviation; $p$: $p$-value from the family-level two-sided paired sign-flip randomization test; Holm $p$: $p$-value adjusted across the 15 features. Bins follow the definition in Appendix~C, from 1 (shallowest) to 5 (deepest); resolution-score bin 1 has no validated layer in any model (value 0), and residual-update bin 5 includes the final step. A negative resolution score difference indicates a weaker signal toward the correct answer in circular systems.}\label{tab:app-features}
\end{table*}

\setcounter{table}{7}
\begin{table*}[t]
\centering\footnotesize
\setlength{\tabcolsep}{3.5pt}
\begin{adjustbox}{max width=\textwidth}
\begin{tabular}{@{}lccccc@{}}
\toprule
 & Frozen & \multicolumn{2}{c}{Training-row scaling} & \multicolumn{2}{c@{}}{Joint holdout}\\
\cmidrule(lr){2-2}\cmidrule(lr){3-4}\cmidrule(l){5-6}
Model & AUC & CV AUC & AUC & AUC & Holm $p$\\
\midrule
Gemma-2-9B & \LomoGem & \SensCVGem & \SensLomoGem & \JointGem & \JointHolmGem\\
Llama-3.2-11B & \LomoLla & \SensCVLla & \SensLomoLla & \JointLla & \JointHolmLla\\
Llama-3.3-70B & \LomoSev & \SensCVSev & \SensLomoSev & \JointSev & \JointHolmSev\\
Qwen3-VL-8B & \LomoQwe & \SensCVQwe & \SensLomoQwe & \JointQwe & \JointHolmQwe\\
\bottomrule
\end{tabular}
\end{adjustbox}
\caption{Post hoc check of classifier preprocessing (exploratory; Appendix~C). Frozen: leave-one-model-out AUC from Table~5, with feature scales estimated using all rows of each model. Training-row scaling: all scales are refit using only the training rows (CV AUC: within-model family-grouped nested cross-validation; AUC: leave-one-model-out, with pooled scales and inner folds grouped by family). Joint holdout: one model and one of the eight family folds are held out together; $p$-values are Holm-adjusted across the four held-out models.}\label{tab:app-sensitivity}
\end{table*}

\FloatBarrier
\end{document}